\documentclass[11pt]{article}
\pdfoutput=1
\usepackage{graphicx}
\usepackage{epsfig}
\usepackage{natbib}
\usepackage{multirow}
\usepackage{media9}    
\usepackage{subfig}
\usepackage{color}
\usepackage{caption}
\usepackage{setspace}
\usepackage{algorithm} 
\usepackage{algorithmic}  
\usepackage{varwidth}
\usepackage{framed}
\usepackage{mathrsfs}
\usepackage{amsmath}
\usepackage{amsfonts}
\usepackage{float}
\usepackage{empheq}
\usepackage{hyperref}
\usepackage{times}
\usepackage{lineno}

\newcommand{\mbf}[1]{\mathbf{#1}}
\newcommand{\mbv}[1]{\mbox{\boldmath$#1$\unboldmath}}

\def\bc{\mathbf{c}}

\def\bs{\mathbf{s}}

\def\bw{\mathbf{w}}
\def\bx{\mathbf{x}}
\def\by{\mathbf{y}}
\def\bz{\mathbf{z}}

\def\bC{\mathbf{C}}

\def\bH{\mathbf{H}}
\def\bI{\mathbf{I}}

\def\bM{\mathbf{M}}

\def\bV{\mathbf{V}}
\def\bU{\mathbf{U}}
\def\bW{\mathbf{W}}
\def\bX{\mathbf{X}}
\def\bY{\mathbf{Y}}

\newcommand{\bftheta}{\mbox{\boldmath $\theta$}}

\newcommand{\bfalpha}{\mbox{\boldmath $\alpha$}}
\newcommand{\bfeta}{\mbox{\boldmath $\eta$}}
\newcommand{\bfbeta}{\mbox{\boldmath $\beta$}}
\newcommand{\bfzeta}{\mbox{\boldmath $\zeta$}}

\newcommand{\bfmu}{\mbox{\boldmath $\mu$}}
\newcommand{\bfnu}{\mbox{\boldmath $\nu$}}
\newcommand{\bfomega}{\mbox{\boldmath $\omega$}}

\newcommand{\bfPhi}{\mbox{\boldmath $\Phi$}}
\newcommand{\bfPsi}{\mbox{\boldmath $\Psi$}}

\newcommand{\bfxi}{\mbox{\boldmath $\xi$}}

\def\var{\textrm{var}}
\def\cov{\textrm{cov}}

\begin{document}

\title{\bf  Comparison of Deep Neural Networks and Deep Hierarchical Models for Spatio-Temporal Data}
\author{Christopher K. Wikle \\ {\small Department of Statistics}\\ {\small University of Missouri, Columbia, MO}}
\date{February 15, 2019}
 \maketitle
\begin{abstract}
Spatio-temporal data are ubiquitous in the agricultural, ecological, and environmental sciences, and their study is important for understanding and predicting a wide variety of processes.   One of the difficulties with modeling spatial processes that change in time is the complexity of the dependence structures that must describe how such a process varies, and the presence of high-dimensional complex datasets and large prediction domains.  It is particularly challenging to specify parameterizations for nonlinear dynamic spatio-temporal models (DSTMs)  that are simultaneously useful scientifically and efficient computationally.  Statisticians have developed deep hierarchical models that can accommodate process complexity as well as the uncertainties in the predictions and inference. However, these models can be expensive and are typically application specific.  On the other hand, the machine learning community has developed alternative ``deep learning'' approaches for nonlinear spatio-temporal modeling.  These models are flexible yet are typically not implemented in a probabilistic framework.  The two paradigms have many things in common and suggest hybrid approaches that can benefit from elements of each framework. This overview paper presents a brief introduction to the deep hierarchical DSTM (DH-DSTM) framework, and deep models in machine learning, culminating with the deep neural DSTM (DN-DSTM).  Recent approaches that combine elements from DH-DSTMs and echo state network DN-DSTMs are presented as illustrations.

 \end{abstract}


\vspace{4mm}

\section{Introduction}
Deep learning is a type of machine learning (ML) that exploits a connected hierarchical set of models to predict or classify elements of complex data sets.  The ML deep learning revolution is relatively recent and primarily associated with neural models such as feedforward neural networks (FNNs), convolutional neural networks (CNNs), recurrent neural networks (RNNs), generative adversarial networks (GANs), or some combination of these neural architectures.  There are remarkable success stories associated with these approaches, such as models that can defeat experts in Go, Chess, or Shogi \citep{silver2016mastering,silver2018general}, and of course, there are failures as well \citep{shalev2017failures}, albeit less publicized.  Statisticians should not be surprised by the success (and failure) of these deep ML methods as we have been using deep hierarchical models (HMs) for years.  By deep hierarchical models, we mean multi-level Bayesian hierarchical models.  Indeed, many of the reasons for success and failure of deep ML and deep HMs are the same.  The primary purpose of this article is to discuss some of these connections in the context of an area of great interest in agriculture, environmental, and ecological statistics -- spatio-temporal modeling, and to show some ways in which deep ML model methodologies can be utilized within a traditional statistical modeling framework.

Spatio-temporal processes are ubiquitous in the environmental sciences.  They describe how spatially-dependent processes change through time, subject to various forcing mechanisms.  An important modeling challenge for such processes concerns how one accounts for interactions between different scales of spatial and temporal variability, internal to the process of interest, as well as how that process interacts with other (exogenous) processes.  Often spatio-temporal processes are quite nonlinear in time, at least at certain time or spatial scales.  It can be difficult to model interactions for such processes in a parsimonious way, although some parametric spatio-temporal statistical models have been used in this context, often by incorporating knowledge about the underlying dynamics of the system of interest \citep[e.g.][]{wikle2001spatiotemporal,wikle2010general}.  Such deep hierarchical dynamic spatio-temporal models (DH-DSTMs) can be quite complex.  Similarly, perhaps the greatest success stories in deep ML methods have been associated with data with complex spatial and temporal dependencies.  In particular, CNN models have been very successful in vision and image processing, and RNN models have exploited the complex temporal dependencies in language processing \citep[see the overviews in][]{goodfellow2016deep,aggarwal2018neural}.  Increasingly, CNN and RNN approaches are being combined to model spatio-temporal processes \citep[e.g.,][]{donahue2015long}.   In this paper, we refer to such hybrid spatio-temporal models as deep neural dynamical spatio-temporal models (DN-DSTMs).

Faced with complex spatio-temporal modeling challenges, how does the environmental statistician decide which paradigm is most appropriate for their problem?
DH-DSTMs and DN-DSTMs can both be challenging to implement -- often requiring a great deal of training data and specialized  computational algorithms.  As discussed in Section \ref{sec:connections}, the two modeling paradigms share common (or, at least similar) solutions to these challenges.   One must also consider how important uncertainty quantification (UQ) is to the problem at hand.   As statisticians, we would like to think that UQ is {\it always} of fundamental importance to what we do, but the reality is that there are situations where one simply needs a prediction or classification and the UQ is secondary at best.  Most DN-DSTM methods do not provide a model-based measure of uncertainty, whereas the DH-DSTM approach is built upon a framework to explicitly capture uncertainty about as many aspects of the problem as possible (data, process, and parameters).  But, DN-DSTM models do have the flexibility to consider non-Markovian feedback mechanisms in time and the influence of specific events in the distant past, whereas DH-DSTMs are typically based on Markovian (i.e., autoregressive) structures.    This suggests we might borrow ideas from both the DH-DSTM and DN-DSTM approaches to develop relatively parsimonious and flexible models that can accommodate real-world complexity and UQ, potentially in a computationally feasible manner.  Perhaps more importantly, in some cases, these methods could be used in situations where one does not have access to tremendous sources of data (either labeled or unlabeled), especially when they are linked together with parsimonious architectures.

Section \ref{sec:overview} provides a concise overview of spatio-temporal modeling in statistics from both the descriptive and dynamic perspective, illustrating the importance of basis-function representations.  This is followed by a brief overview of deep modeling and the DH-DSTM statistical perspective in Section \ref{sec:deepmodels}.  Section \ref{sec:deepNN} then gives a brief overview of deep models in machine learning and issues associated with their implementation, including deep feedforward NNs (DNNs), CNNs, RNNs, and DN-DSTMs.  Section \ref{sec:linking} then reviews some recent approaches for linking the DH-DSTM and DN-DSTM frameworks.  A concluding discussion is presented in Section \ref{sec:discussion}.


\section{A Brief Overview of Spatio-Temporal Modeling}\label{sec:overview}

In statistics, we have typically been interested in spatio-temporal models that follow the general form of an observation model and a model for a spatio-temporal latent process \citep[e.g.][]{cressie2011statistics,wikle2019spatio}:
\begin{eqnarray}
[\mbox{observations} \; & | & \; \mbox{latent process} \;\; \mbox{and} \;\; \mbox{obs/sampling error}] \label{eq:obs}\\
  \mbox{latent process} \; & = & \; \mbox{``fixed effects''} \;\; + \;\; \mbox{dependent random process}, \label{eq:proc}
\end{eqnarray}
where $[ \;]$ denotes a generic distribution, $|$ denotes conditioning, and each component of the model is indexed in space and time.  More formally, assume we are interested in a latent (unobserved) spatio-temporal process $\{Y(\bs;t): \bs \in D_s, t \in D_t\}$ where $\bs$ is a spatial location in domain $D_s$ (a subset of $d$-dimensional real space) and $t$ is a time index in temporal domain $D_t$ (along the one-dimensional real line).  We then have observations $\{z(\bs_{ij};t_j)\}$ for spatial locations $\{\bs_{ij}: i=1,\ldots,m_j\}$ and times $\{t_j: j=1,\ldots,T\}$.

A common example of (\ref{eq:obs}) for Gaussian spatio-temporal observations is given by
\begin{equation}
z(\bs_{ij};t_j) =  Y(\bs_{ij};t) + \epsilon(\bs_{ij};t_j),  \label{eq:Zdata}
\end{equation}
where $\epsilon(\bs_{ij};t_j) \; \sim iid \; Gau(0,\sigma^2_\epsilon)$ is the observation error process.  The latent Gaussian spatio-temporal process (\ref{eq:proc}) can be represented as
\begin{equation}
Y(\bs;t) = \mu(\bs;t) + \eta(\bs;t) \label{eq:Yproc},
\end{equation}
where $\mu(\bs;t)$ is a spatio-temporal mean function, and $\eta(\bs;t)$ is a mean zero Gaussian process (GP) with covariance function, say $c_\eta(\eta(\bs;t),\eta(\bs',t')) \equiv \cov(\eta(\bs;t),\eta(\bs';t'))$.  Then, $Y(\bs;t)$ is also a GP with mean function $\mu(\bs;t)$ and covariance function $c_\eta(\cdot,\cdot)$.  Recall, a GP is a distribution over {\it functions} that is fully specified by a mean function and covariance function defined over the spatio-temporal domain of interest (e.g., $D_s \times D_t$).  GPs have the very useful property that all of their finite-dimensional distributions are Gaussian (i.e., normal). 

Now, say we are interested in predicting the latent process at location $(\bs_0;t_0)$ given the $m = \sum_j m_j$-dimensional observation vector $\bz \equiv \{z(\bs_{ij};t_j)\}$.  The spatio-temporal (universal) kriging optimal predictor is the linear predictor $\widehat{Y}(\bs_0;t_0)$ that minimizes the mean squared prediction error, $E(Y(\bs_0;t_0) - \widehat{Y}(\bs_0;t_0))^2$:
\begin{equation}
\widehat{Y}(\bs_0;t_0) = \bx(\bs_0;t_0)' \widehat{\bfbeta}_{gls} + \bc_0' \bC_z^{-1} (\bz - \bX \widehat{\bfbeta}_{gls}),
\label{eq:ukYgls}
\end{equation}
where $\bx(\bs_0;t_0)$ is a $p$-vector of covariates known at all observation locations and at location $(\bs_0;t_0)$, $\bfbeta$ is the associated parameter vector, $\bX$ is the $m \times p$ matrix of covariates at observation locations, $\bC_z \equiv \var(\bz)$ is an $m \times m$ covariance matrix, $\bc_0 \equiv  c_\eta(\bz,Y(\bs_0;t_0)$ is the $m \times 1$ covariance vector between observation locations and the prediction location, and the generalized-least-squares (gls) estimator of $\bfbeta$ in (\ref{eq:ukYgls}) is given by $\widehat{\bfbeta}_{gls} \equiv (\bX' \bC_z^{-1} \bX)^{-1} \bX' \bC_z^{-1} \bz.$  Note that $\bC_z = \bC_y + \sigma^2_\epsilon \bI = \bC_\eta + \sigma^2_\epsilon \bI$.  The associated spatio-temporal kriging variance is given by $\sigma^2_{Y}(\bs_0;t_0) = c_{0,0} - \bc_0' \bC_z^{-1} \bc_0 + \kappa$,
where $c_{0,0} \equiv \var(Y(\bs_0;t_0))$ and
$\kappa$ represents the uncertainty brought to the prediction due to the estimation of $\bfbeta$ \citep[e.g.,][]{wikle2019spatio}.  It is straight forward to modify these formulas to obtain predictions for many locations at once, and the approach can be extended to non-Gaussian data models as well, but without a closed form solution \citep[e.g., see][]{cressie2011statistics}. 

\begin{figure}
\begin{center}
\includegraphics[width=3in]{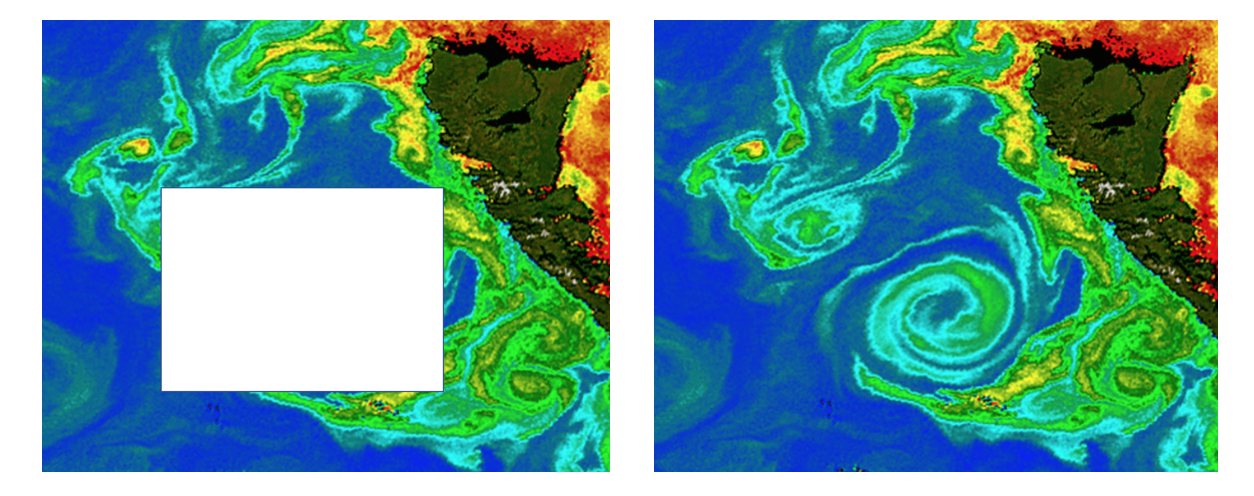} 
\caption{Ocean color images from the SeaWiFS satellite -- note that ocean color is a proxy for phytoplankton primary productivity in the ocean.  The left panel shows a schematic box representing missing observations as one often experiences due to cloud cover.  The right panel shows that there is a mesoscale eddy (a medium-spatial scale highly nonlinear circulation feature) in that region.  This illustrates the challenge of trying to use traditional interpolation-based spatial or temporal prediction methods for complex processes.}
\label{fig:oceancolorexample}
\end{center}
\end{figure}

This approach to spatio-temporal modeling is {\it descriptive} (marginal) in that it only relies on the first and second moments of the latent process $\{Y(\bs;t)\}$.  In the spatio-temporal context this is quite useful when one does not have a great deal of knowledge about the underlying process and only needs to specify a plausible spatio-temporal covariance structure (and a spatio-temporal trend) and can rely in some sense on ``Tobler's law'' that nearby things in space (and time) are more related than distant things \citep{tobler1970computer}.   However, this can be challenging for complex processes as it is difficult to specify valid covariance functions that are realistic in many situations where Tobler's law might not hold (e.g., eddy dynamics, density-dependent growth, etc.).  In addition, such second-order moment-based approaches are limiting for nonlinear and non-Gaussian processes.  Practically, as shown in Figure \ref{fig:oceancolorexample}, these limitations are most noticeable in situations where one is forecasting multiple time steps into the future and/or must fill in large gaps in the spatio-temporal domain of interest.  

\subsection{Dynamic Spatio-Temporal Models (DSTMs)}

The dynamical approach to spatio-temporal process modeling in statistics is based on the idea of conditioning the spatial process at the current time on the recent past (i.e., a Markov assumption).  The model is primarily concerned with specifying the evolution of the spatial field through time.  This specification of the evolution of the spatial process describes the etiology of the environmental process.  Such specifications have traditionally worked well when one has some underlying knowledge about the process of interest to help with estimation of the transition operator that controls the evolution  \citep[e.g.,][]{wikle2010general}.  These models are typically most effective when forecasting multiple time steps in the future and/or predicting across large regions of space in which there are no observations.  

The data model in a general DSTM can be written
\begin{equation}
{z}_t(\cdot)  =  {\cal{H}}({Y}_t(\cdot), {\mbv \theta}_{d,t}, {\epsilon}_t(\cdot)), \;\;\; t=1,\ldots,T,
\label{eq:dyn_Z}
\end{equation}
where ${z}_t(\cdot)$ corresponds to the data at time $t$, ${Y}_t(\cdot)$ the corresponding latent process of interest, with a linear or nonlinear mapping function, ${\cal{H}}(\cdot)$, that relates the data to the latent process. The data model error is given by ${ \epsilon}_t(\cdot)$, and data model parameters are represented by ${\mbv \theta}_{d,t}$. These parameters may vary spatially and/or temporally in general.  An important assumption that is present here, as well as in the descriptive model presented above, is that the data ${z}_t(\cdot)$ are independent in time when conditioned on the true process, ${Y}_t(\cdot)$, and parameters ${\mbv \theta}_{d,t}$.  (Note, as is customary  in dynamic models, we represent the time index as a subscript here.)

The most important component of the DSTM is the dynamic process model.  One can simplify this by making use of conditional independence through Markov assumptions (e.g., conditioned on the recent past, the process is independent of the process in the more distant past).  For example, a first-order Markov process can be written
\begin{equation}
{Y}_t(\cdot)  =  {\cal M}({Y}_{t-1}(\cdot), {\mbv \theta}_{p,t}, { \eta}_t(\cdot)), \;\;\; t=1,2,\ldots,
\label{eq:Yprocess}
\end{equation}
where ${\cal M}(\cdot)$ is the evolution operator (linear or nonlinear),  ${\eta}_t(\cdot)$ is the noise (error) process, and ${\mbv \theta}_{p,t}$ are process model parameters that may vary with time and/or space.  Note that here we are assuming that time is discrete and equally spaced (although this can be relaxed).  As an example, a linear evolution equation would be written, $\bY_t = \bM \bY_{t-1} + \bfeta_t$, where $\bfeta_t \sim Gau({\mbf 0},\bC_\eta)$, $\bY_t$ is an $n \times 1$ vector corresponding to spatial locations, $\bM$ is a transition matrix of dimension $n \times n$,  and $\bC_\eta$ is the $n \times n$ innovation error covariance matrix (a spatial covariance matrix in this case).  Typically, one would also specify a distribution for the initial state,  $[{Y}_0(\cdot) | {\mbv \theta}_{p,0}]$.  

Finally, one either estimates the parameters in (\ref{eq:dyn_Z}) and (\ref{eq:Yprocess}) directly, or assigns them distributions.  As discussed below, an important part of the DH-DSTM framework is modeling these parameters as processes (e.g., spatially or temporally varying, and/or allowing them to depend on auxiliary covariate information, etc.).

\subsection{Basis Function Representation}

Both the descriptive and dynamic approaches to spatio-temporal modeling suffer from a curse-of-dimensionality.  In the descriptive case, we need to be able to efficiently calculate the inverse $\bC_z^{-1}$, and in the dynamic case, we need to be able to estimate the parameters in the transition operator (e.g., the transition matrix $\bM$ in the linear case).  This is challenging if the number of spatial locations (data and/or prediction) is large.  There are a number of ways in which these issues can be mitigated  \citep[e.g., see the overview,][in the context of spatial models]{heaton2018case}, but a common approach to both is to consider basis function representations.  

Consider expanding the spatio-temporal process in a finite-dimensional basis expansion:
\begin{equation}
Y(\bs;t) = \bx(\bs;t)' {\mbv \beta} + \sum_{i=1}^{n_\alpha} \phi_i(\bs) \alpha_i(t) + \nu(\bs;t),
\label{eq:Ybasis}
\end{equation}
where $\{\phi_i(\bs):i=1,\ldots,n_\alpha\}$ are basis functions, $\{\alpha_i(t):i=1,\ldots,n_\alpha\}$ are the associated random expansion coefficients, and $\nu(\bs;t)$ is a relatively simple spatio-temporal process sometimes needed to represent left-over fine-scale spatio-temporal random variation.  Note that we could consider basis functions that are indexed in space and time, or just time \citep[e.g., see][]{wikle2019spatio}.  

Of course, there is a well-known connection between covariance functions, basis functions, and kernels in the context of Mercer's theorem and the Karhunen-Lo\'eve decomposition for GPs \citep[e.g., see][]{rasmussen2006gaussian}.  But, to see the practical utility of this representation, one need only note that they allow us to to build complexity through marginalization in a computationally efficient manner.  For example, recall from linear mixed model theory that we can write (in vector/matrix form) the conditional model
$$
\bY | \bfalpha \sim Gau(\bX \bfbeta + \bfPhi \bfalpha, \bC_\nu),
$$
$$
\bfalpha \sim Gau({\mbf 0},\bC_\alpha).
$$
Then, integrating (marginalizing) out the random effects $\bfalpha$ induces dependence:
$$
\bY \sim Gau(\bX \bfbeta, \bfPhi \bC_\alpha \bfPhi' + \bC_\nu).
$$
That is, we have constructed the marginal covariance matrix through the known basis functions and the dependence in the random effects: $\bC_y = \bfPhi \bC_\alpha \bfPhi' + \bC_\nu$.  

In this context, the main spatio-temporal dependence structure comes from either $\bC_\alpha$ in the descriptive case, or $\bfalpha_t = \bM_\alpha \bfalpha_{t-1} + \bfeta_t$ in the dynamic case.  Then, the computational advantage of basis functions comes when one recognizes that $\{\bfalpha_t\}$ is simpler than $\{Y(\bs;t)\}$ so that $\bC_\alpha^{-1}$ and/or $\bM_\alpha$ are easy to obtain.  This occurs when one is working with a low-rank system (i.e., $n_\alpha \ll n$) or when there are efficient algorithms for manipulating the basis functions and/or random effects \cite[e.g., see][]{cressie2011statistics}.  Basis function approaches can be quite useful for spatio-temporal modeling, but there are still many situations that require more complicated process descriptions on the random effects.  This is best considered from a hierarchical modeling perspective.

\section{Multi-Level (Deep) Hierarchical Models}\label{sec:deepmodels}

What are deep models?  Although there is probably no universally agreed upon answer, one view is that a deep model is structured so that the response (output) is given by a sequence of linked (telescoping) models:
$${
\mbox{Response (Output)} \longleftarrow m_1 \longleftarrow m_2 \longleftarrow \cdots   \longleftarrow  m_L (\longleftarrow \mbox{Input}),
}$$
where $m_\ell$ corresponds to the $\ell$th model.   In the statistics, this is perhaps best represented by the Bayesian hierarchical modeling framework \citep[e.g., see][]{gelman2006data,gelman2013bayesian}, in which case the input is {\it not} included at the deep end of the model, but can be in any stage, or at the top.  In particular, in the context of environmental statistics, the hierarchical modeling paradigm of \citet{berliner1996hierarchical}, \citet{wikle1998hierarchical}, and \citet{cressie2011statistics} considers the following general distributions/models:

$$\mbox{Data Models:} \;\;  [data \; | \; process, data \;
parameters]$$
$$\mbox{Process Models:} \;\;[process \; | \; process \;
parameters]$$
$$\mbox{Parameter Models:} \;\; [data \; 
and \; process \; parameters].$$
For inference and prediction one evaluates the posterior distribution:
$$
\mbox{Posterior:} \;\; [process, \; parameters \; | \; data ],
$$
which is proportional to the product of the data, process, and parameter distributions given above.  Typically, there are multiple sub-stages for each level, which adds to the model depth. The key to the \citet{berliner1996hierarchical} HM paradigm (which, unfortunately, is often ignored) is that {\it one avoids modeling second-order structure as much as possible.}  That is, one puts the modeling effort into the conditional mean to build dependence (complexity) through marginalization (as with the basis function illustration discussed above).  So, these are linked conditional models  and very much top down in the sense that inputs are usually closer to the top (data) level, although they can enter at any level in principle.   The next section illustrates the general DH-DSTM deep model for complex spatio-temporal modeling.

\subsection{Deep Hierarchical Dynamical Spatio-Temporal Modesl (DH-DSTMs)}\label{sec:H-DSTMs}

Here we outline a prototypical DH-DSTM.  For simplicity, and to compare to the deep ML models in Section \ref{sec:deepNN}, this model is presented in the context of discrete time and space, although time and/or space can be considered continuous more generally.  For $t=1,\ldots,T$,
\begin{eqnarray}
{\mbox{Data Model:       }} & \; &  \;\;\;  \bz_t | \bY_t, \bftheta_h  \sim \;  {\cal D}(\bH_t \bY_t; \bftheta_h), 
\label{eq:DSTM_datamodel} \\
{\mbox{ Conditional Mean:     }} & \;&   f(\bY_t)  = \bfmu_t + \bfPhi \bfalpha_t + \bfnu_t,   
\label{eq:DSTM_condmean} \\
{\mbox{Process Mean:     }} & \; &  \bfmu_t  = {\mbf W}_t  \bftheta_{\mu} + {\mbv \gamma}_t,
\label{eq:DSTM_mean} \\
{\mbox{ Dynamic Process:     }} & \; & { \bfalpha_t   = g(\bfalpha_{t-\tau},  \bx_{t-\tau}; \bftheta_\alpha; \bfeta_t)},
\label{eq:DSTM_dynproc} \\
{\mbox{ ``Residual'' Process:     }} & \; & { [\bfnu_t | \bftheta_{\nu}]},
\label{eq:DSTM_residST} \\
{\mbox{ Regularization Priors:     }} &\; & { [\bftheta_\alpha | {\bfzeta}]}, \nonumber \\
{\mbox{Parameters:     }}  &\;&  {[\bftheta_h, \bftheta_{\nu}, \bftheta_{\mu}, \bfzeta]}.  \nonumber
\end{eqnarray}

The data model (\ref{eq:DSTM_datamodel}) specifies the distribution for $\bz_t$, which is a spatially-referenced data vector at time $t$. Specifically,  ${\cal D}( \cdot)$ is some generic distribution (e.g., exponential family; this is problem specific), $\bH_t$ is a mapping matrix that maps the latent process locations to the data locations, $\bY_t$ is the spatially referenced latent process vector at time $t$,  and $\bftheta_h$ are data model parameters. The important assumptions in this data model are that the observation vectors are considered to be independent conditioned on the latent process, and the observation error structure is relatively simple (i.e., independent) since most of the dependence is attributed to the latent process.  Note also that multiple data (input) sources can easily be accommodated as in the general \citet{berliner1996hierarchical} framework.  

The conditional mean (\ref{eq:DSTM_condmean}) specifies a transformation (link function) $f(\cdot)$,  where $\bfmu_t$ is a time-varying spatial ``trend'' (note, this can depend on inputs, {$\bx_t$}), $\bfPhi$ is a matrix of spatial basis functions (providing dimension reduction), $\bfalpha_t$ is a latent dynamical random process ($n_\alpha \ll n_y$), and $\bfnu_t$ is a non-dynamic spatio-temporal random process (described below).  The most important assumption of this portion of the model is that the latent dynamical process $\{\bfalpha_t\}$ is low dimensional.

The process mean is given in (\ref{eq:DSTM_mean}), where ${\mbf W}_t$ contains covariate inputs to accommodate trends, biases, seasonality, etc., $\bftheta_\mu$ are the associated parameters, and ${\mbv \gamma}_t$ is an error process (typically, Gaussian).  Note that more flexible functions of the covariates can be considered here (i.e., as in generalized additive models) if necessary, but most of the complex structure in the data is due to the $\bfalpha_t$ term described below.  Note also that ${\mbv \gamma}_t$ is assumed to have mean zero and and is typically assumed to be independent in time and space.

The dynamic portion of the model is given by (\ref{eq:DSTM_dynproc}), where $g(\cdot)$ is the evolution operator (potentially nonlinear in ${ \bfalpha_{t-\tau}}$ and inputs ${ \bx_{t-\tau}}$), $\bftheta_\alpha$ are parameters, and $\bfeta_t$ is a noise process (typically assumed to be Gaussian and mean zero, with dependence structure that depends on the specific problem).  This model is arguably the most important part of the DH-DSTM.  It is typically highly parameterized and can, if information is available, be formulated in terms of a mechanistic model, or at least is motivated by such models.  Regardless, it is crucial that this dynamical model allow for interactions in the elements of $\bfalpha_t$ through time \citep[see the discussion in][Chapter 5]{wikle2019spatio}.  As an example, consider the general quadratic nonlinear (GQN) model of \citet{wikle2010general}: 
\begin{equation}
{ \alpha_t(i) = \sum_{j=1}^p \theta^L_{i,j} \; \alpha_{t-\tau}(j) + \sum_{k=1}^p \sum_{\ell = 1}^k \theta^Q_{i,k \ell} \; \alpha_{t-\tau}(k)g(\alpha_{t-\tau}(\ell),\bx_t;\bftheta_g) + \eta_{t}(i)},
\label{eq:GQN}
\end{equation}
where the evolution of an individual $\bfalpha_t$ component is controlled by linear interactions (the first term on the right-hand side (RHS) with parameters $\theta^L$) and quadratic interactions (the second term on the RHS with parameters $\theta^Q$), plus a noise term.  The function $g(\cdot;\cdot)$ is a transformation function that is used to limit the explosive growth induced by the non-linear interactions.  This model is motivated by a wide variety of processes in the physical and biological sciences \citep[see][]{wikle2010general}  and can be quite flexible.  However, this model is severely over-parameterized with $O(p^3)$ parameters, and it requires either science-based hard thresholding or regularization/sparcity on teh parameters for practical implementation.

The residual spatio-temporal process is given in (\ref{eq:DSTM_residST}), where the distribution is determined by the specific problem.  For example, a useful parameterization is to assume another basis expansion such as $\bfnu_t = \bfPsi \bfomega_t + \bfxi_t,$
where $\bfPsi$ is a spatial basis function matrix, $\bfomega_t$ are expansion coefficients, and $\bfxi_t$ is a simple error process \citep[e.g.,][]{wikle2001spatiotemporal}.  The assumption here is that the complex spatio-temporal dynamics are being captured by $\bfalpha_t$, so $\bfomega_t$ would have a simple distribution (e.g., Gaussian with perhaps simple time dependence but independent in ``$\omega$ space''), and $\bfxi_t$ would be independent in time and space.  

As discussed above, the dynamic model for $\bfalpha_t$ is likely over-parameterized and often requires regularization.  Any of the common approaches for regularization in the context of Bayesian models could be used here (e.g., stochastic search variable selection, spike-and-slab, horseshoe priors, etc.; \citet[e.g., see][]{fan2010selective}).  Lastly, we require distributions or fixed values for the remaining parameters. Importantly, in the deep DH-DSTM, these parameters may themselves be ``processes'' (spatial or temporal) and can include dependence on various exogenous input variables.  Implementation of such a deep/complex Bayesian model is typically through problem-specific MCMC algorithms, although there have been recent attempts to consider fairly complex DSTMs in a variational Bayesian context \cite[e.g.,][]{quiroz2018gaussian}.   In general, MCMC implementations can be time consuming and require significant amounts of data, prior information, and computing resources to be successful.  

\subsection{DH-DTSM Example: Ocean Color} 

\begin{figure}
\begin{center}
\includegraphics[width=0.22\textwidth,keepaspectratio=true]{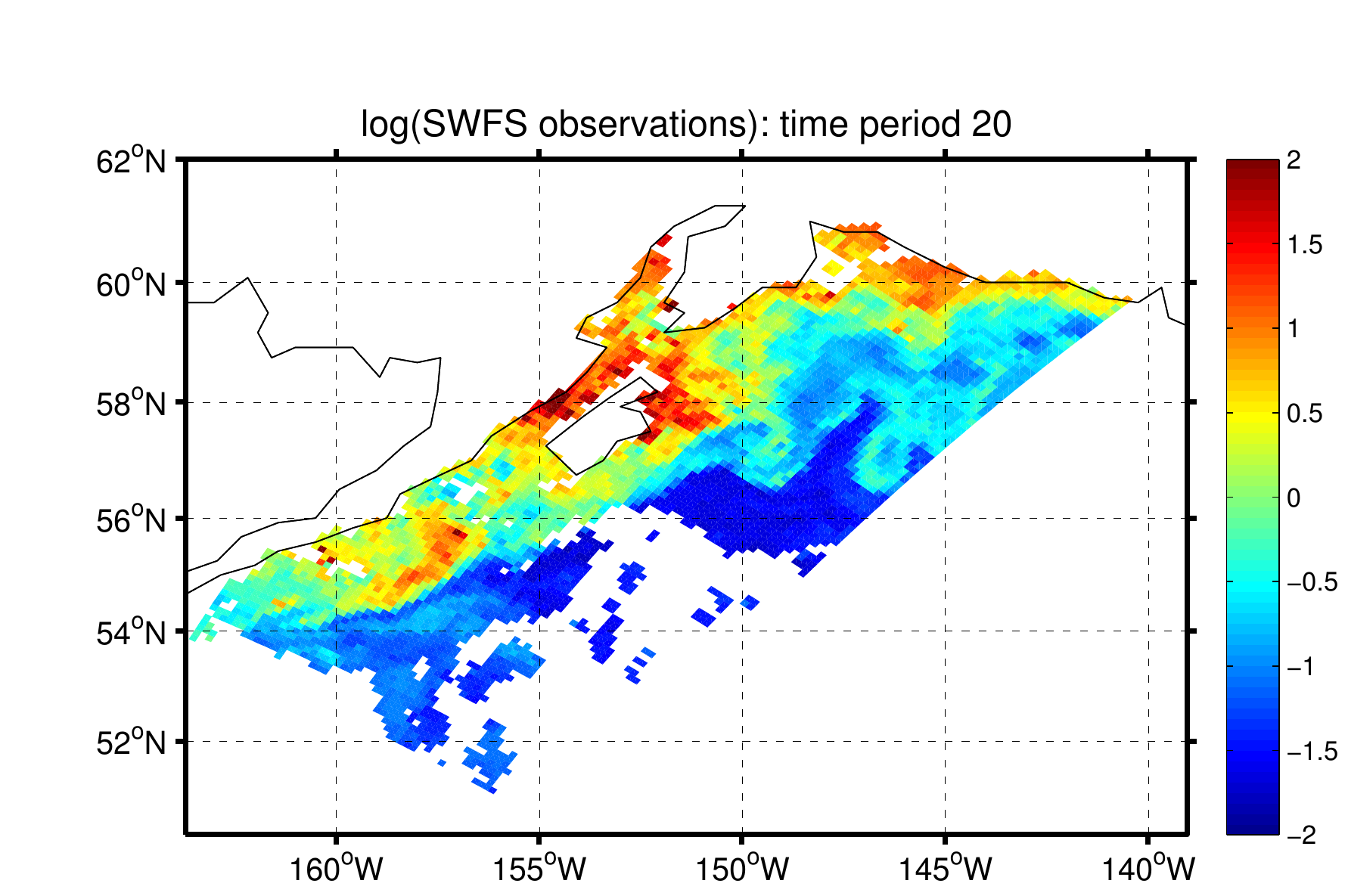}
\includegraphics[width=0.22\textwidth,keepaspectratio=true]{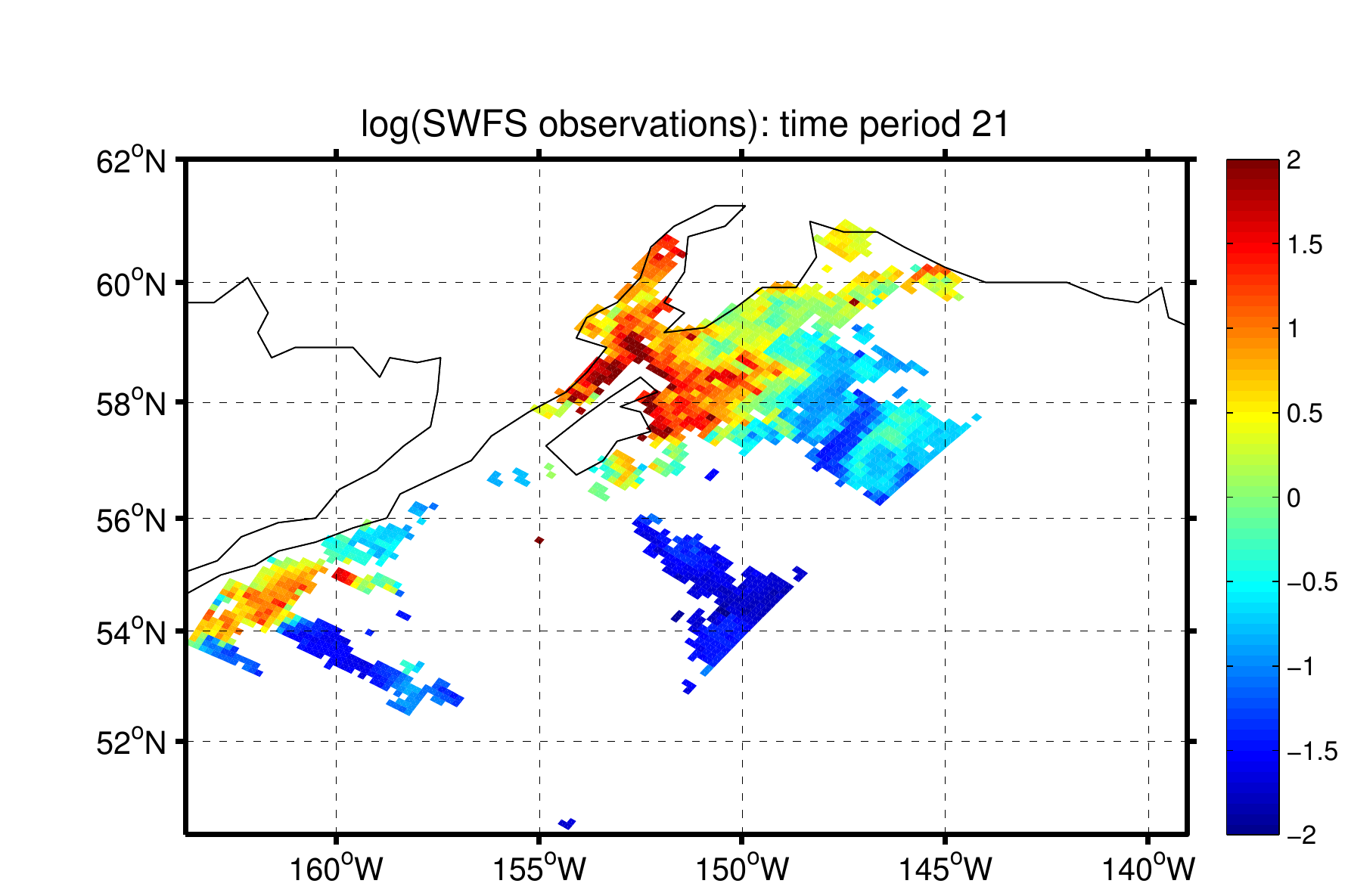}
\includegraphics[width=0.22\textwidth,keepaspectratio=true]{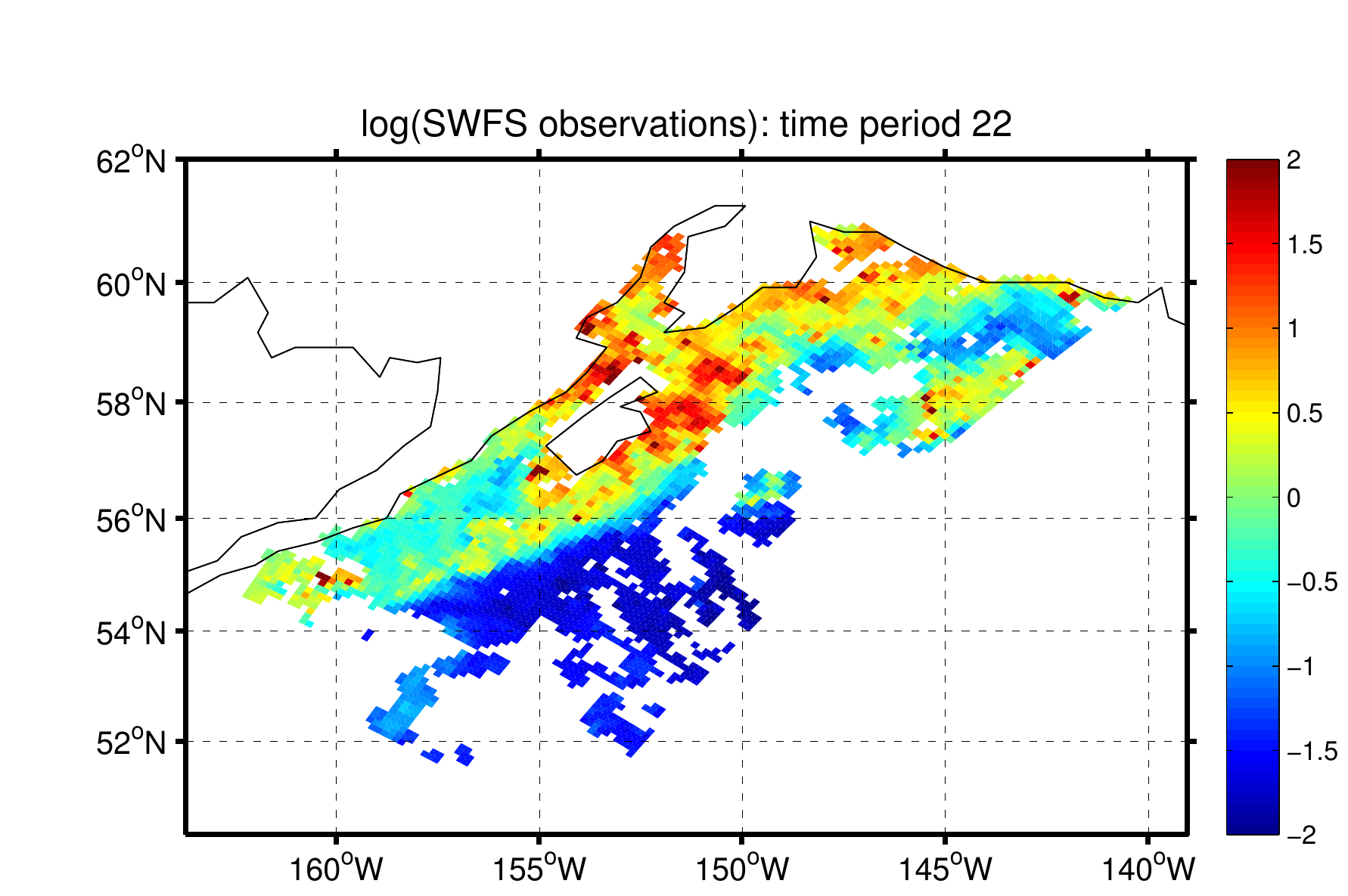}\\
\includegraphics[width=0.22\textwidth,keepaspectratio=true]{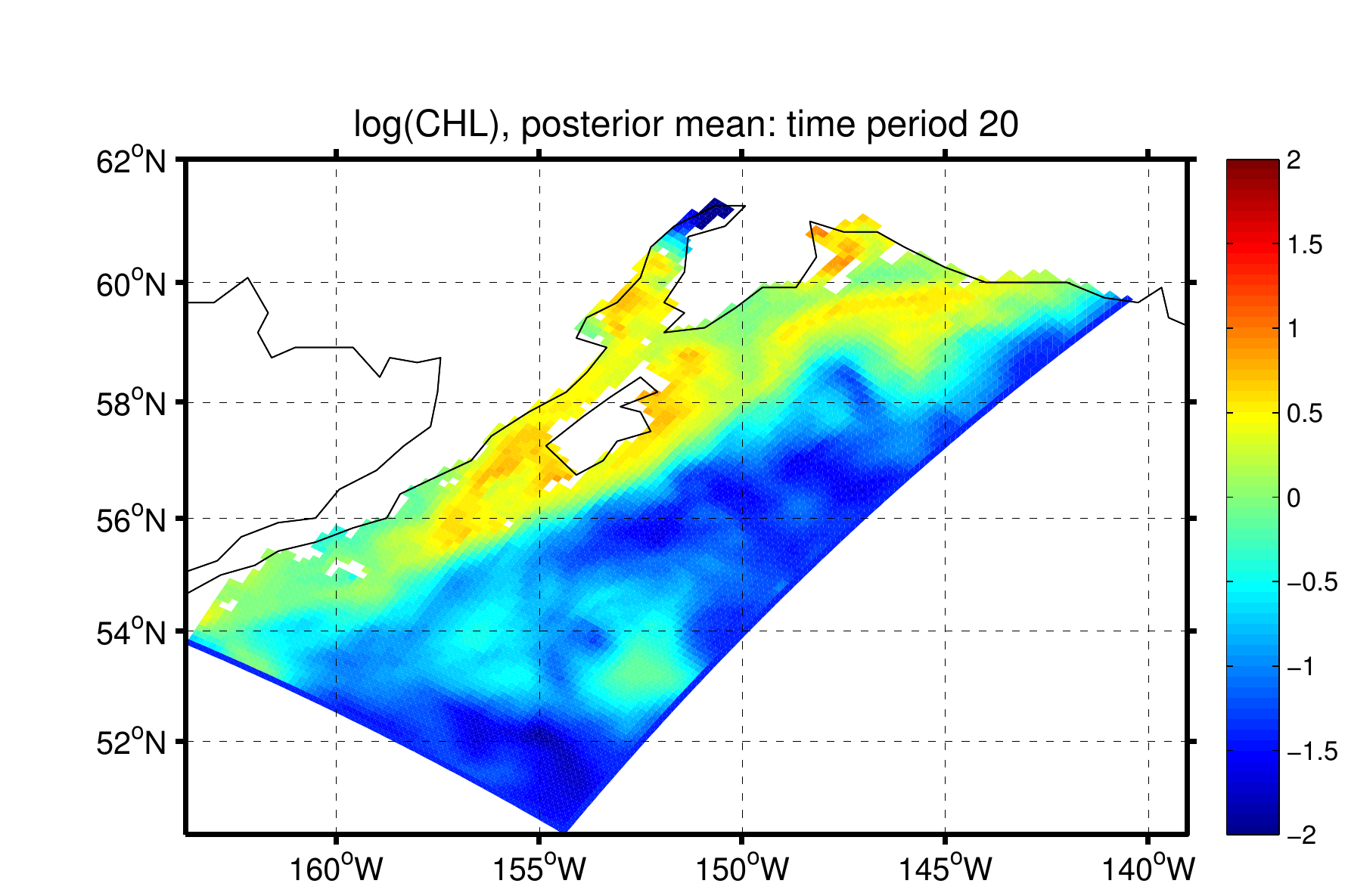}
\includegraphics[width=0.22\textwidth,keepaspectratio=true]{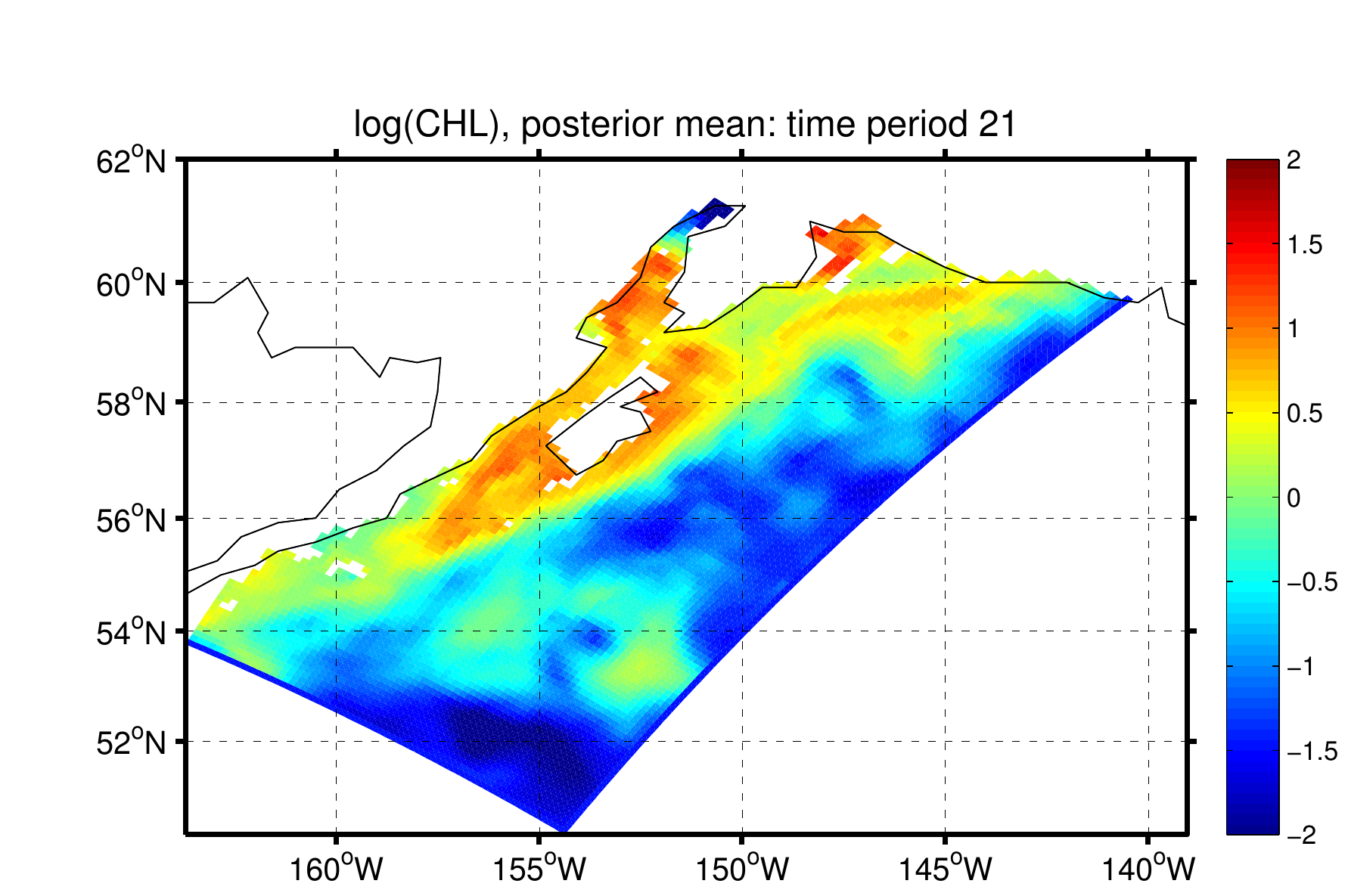}
\includegraphics[width=0.22\textwidth,keepaspectratio=true]{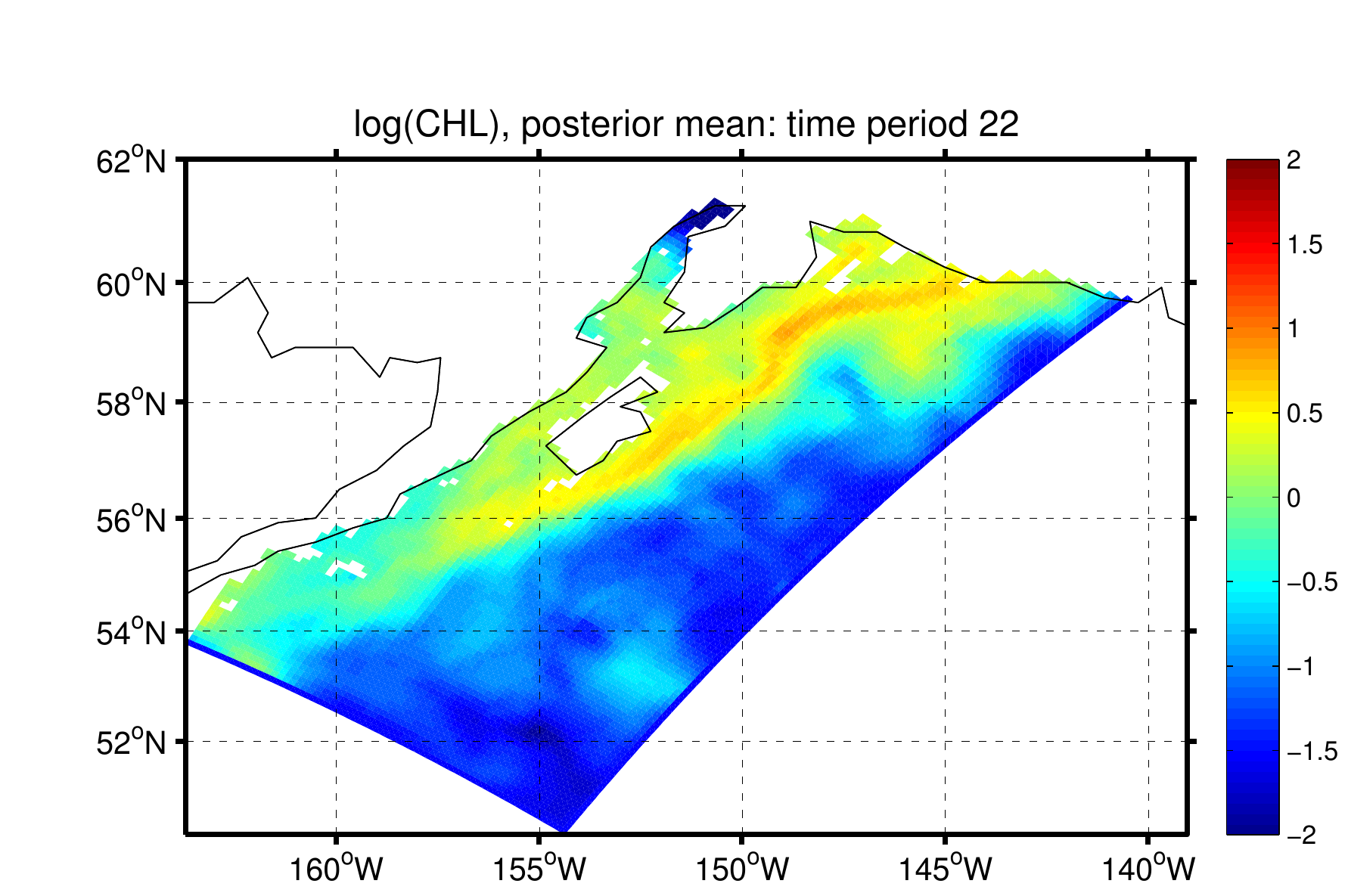}\\
\includegraphics[width=0.22\textwidth,keepaspectratio=true]{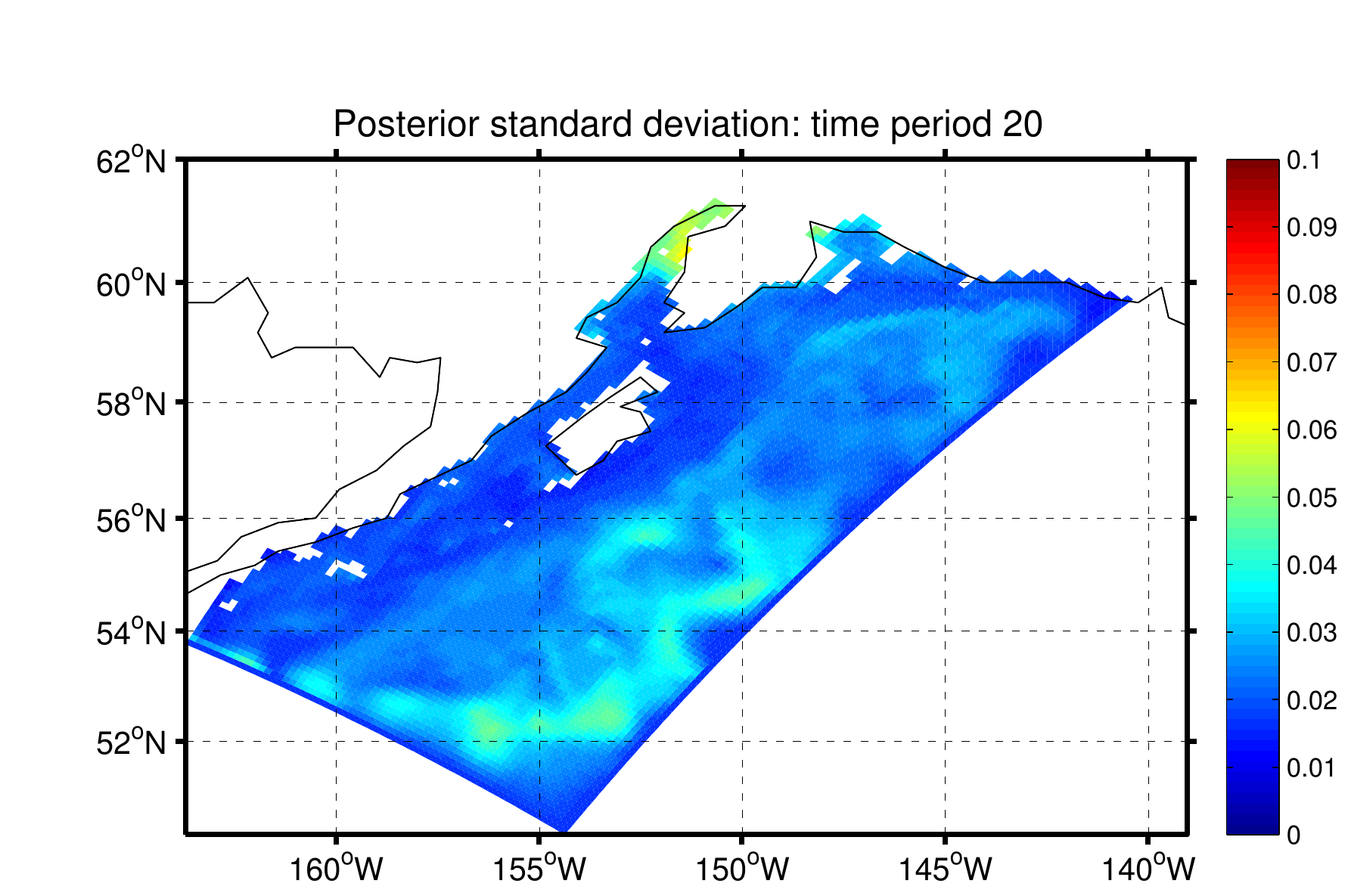}
\includegraphics[width=0.22\textwidth,keepaspectratio=true]{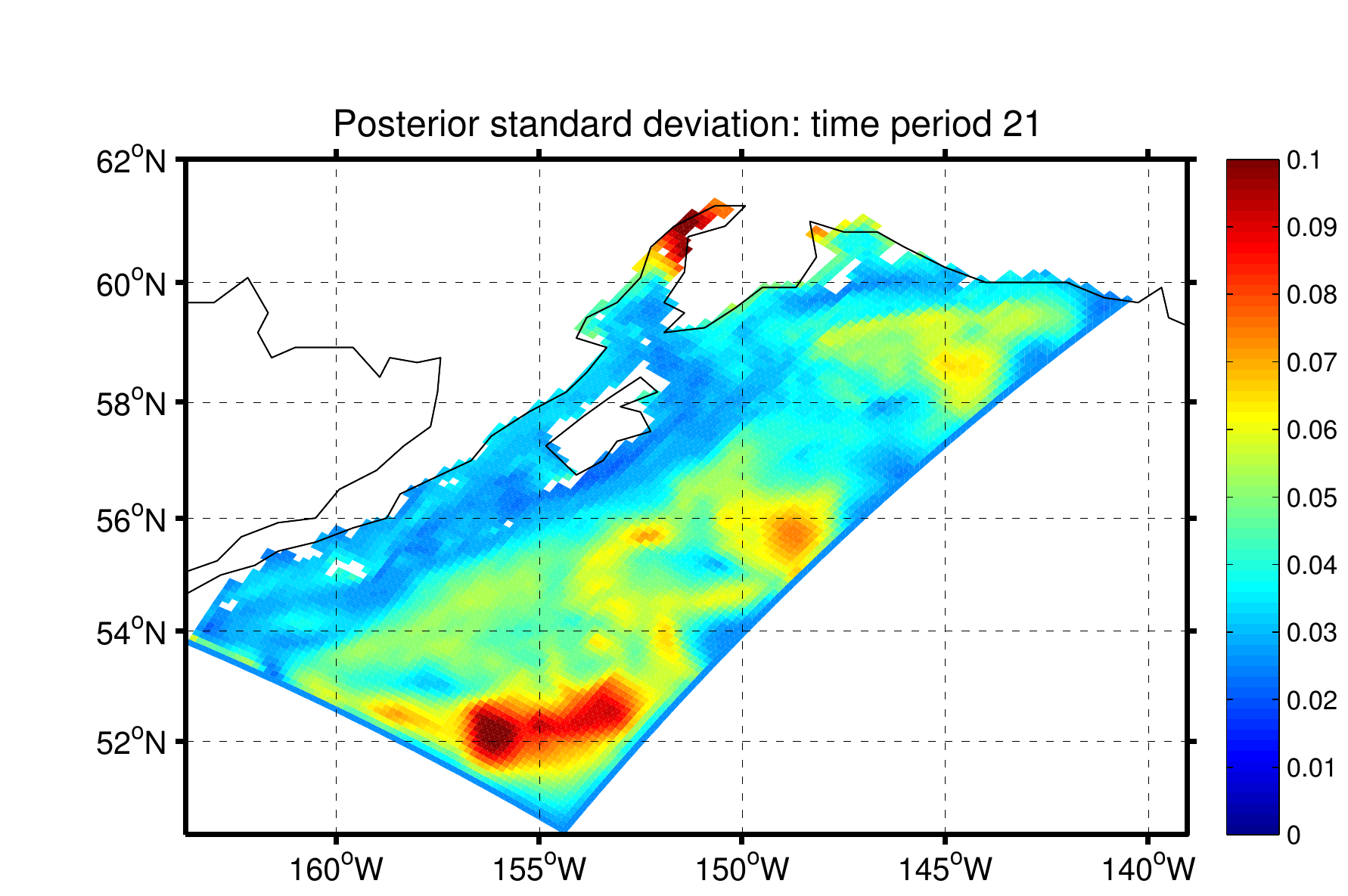}
\includegraphics[width=0.22\textwidth,keepaspectratio=true]{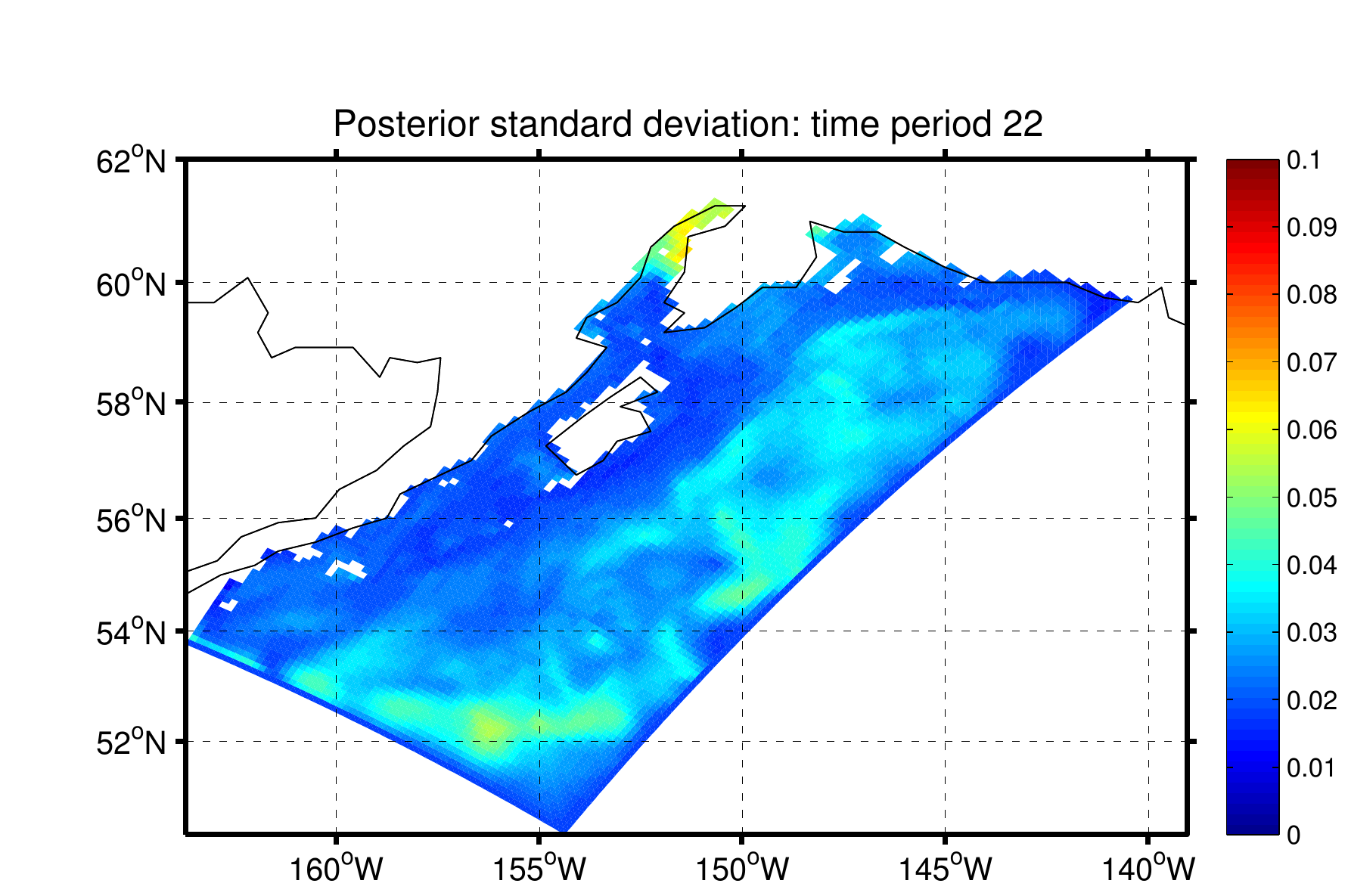}
\caption{Plots of log-transformed SeaWiFS ocean color observations (top row), DH-DSTM posterior mean (second row), and DH-DSTM posterior standard deviation (third row), for three eight-day time periods: June 2, 2002 to June 9, 2002 (left column), June 10, 2002 to June 17, 2002 (center column), and June 18, 2002 to June 25, 2002 (right column).}
\label{fig:seawifs_hdstm}
\end{center}
\end{figure}

\citet{leeds2014emulator} used an DH-DSTM model to perform spatio-temporal prediction to fill gaps in SeaWiFS ocean color observations similar to the issue shown in Figure \ref{fig:oceancolorexample}.  They considered a multivariate model that, in addition to the SeaWiFS observations, included sea surface height (SSH) and sea surface temperature (SST) output from the Regional Ocean Model System (ROMS) that was coupled with a biogeochemical model for the lower trophic ecosystem. They implemented a reduced-dimension GQN process model similar to (\ref{eq:GQN}) as an emulator of the ROMS model (e.g., the ROMS model output was used to train prior distributions for the GQN model -- analogous to ML pre-training described below).  Details can be found in \citet{leeds2014emulator}. As shown in Figure \ref{fig:seawifs_hdstm}, the model was able to predict an eddy in the phytoplankton field despite the fact that the cloud cover in the coastal Gulf of Alaska region left persistent gaps in the SeaWiFS data.  Importantly, the probabilistic nature of the model produces uncertainty measures that suggest that the biggest uncertainty is not that there was an eddy in this area, but rather its precise location. 

\section{Deep Neural Models}\label{sec:deepNN}

The development and application of deep neural models has advanced rapidly over the last decade.  Broad overviews can be found in textbooks such as \citet{goodfellow2016deep} and \citet{aggarwal2018neural}.   The purpose of this section is not to give such a comprehensive treatment, but rather a brief overview to facilitate the connection to DSTMs.  We describe simple feedforward neural networks (NNs), deep feedforward NNs (DNNs), convolutional neural networks (CNNs), and recurrent neural networks (RNNs).  This then provides the background to discuss deep ML models for spatio-temporal data, which we call deep neural DSTMs (DN-DSTMs).

\subsection{Neural Networks}

We start with a very  simple neural network called a  {\it single hidden layer feedforward network}  or {\it single layer perceptron}.  Assume we have a $p$-dimensional input vector $\bx$, and response (output) vector $\bz$, which is $m$-dimensional (but note, $m=1$ in most nonlinear regression and binary classification problems).   We now seek a nonlinear model for the responses given a transformation of the inputs through a ``hidden layer'' given by
\begin{equation}
y_j = g(\sum_{i=0}^p w_{ji} x_i), \;\;\;\; j=1,\ldots,J,
\label{eq:NNactivation}
\end{equation}
where $y_j$ is the hidden variable,  $\{w_{ji}\}$ are the weights (parameters) in which $w_{j0}$ are the bias or offset (intercept) parameters (note, $x_0\equiv 1$), and $g( \cdot)$ is an {\it activation function} (e.g., a hyperbolic tangent, radial basis function, rectified linear unit, softmax, etc.).  The ``output layer'' is then given by:
$$
z_k = g_o( \sum_{j=0}^J v_{kj} y_j), \;\;\; k=1,\ldots,m,
$$
where $g_o(\cdot)$ is an activation function (which may be the identity function), $y_0 \equiv 1$, and $\{v_{kj}\}$ are output weights, including an offset.  One can think of the hidden layer transformation as a basis expansion of the inputs, in which case we can simply write the model as:
$$
z_k({\mbf x};{\mbf W},{\mbf V}) = g_o\left(\sum_{j=0}^J v_{kj} \; g(\sum_{i=0}^p w_{ji} x_i)\right),
$$
where $\bW = \{w_{ji}\}$, and $\bV = \{v_{kj}\}$ and we note that there is no explicit error term in this model.

As with traditional nonlinear regression, to estimate the parameters in such a model (i.e., ``train the network'') we select an objective function in terms  of $\{{\bW},{\bV}\}$ (e.g., squared error, cross-entropy) and then typically use a gradient-based approach to obtain parameter estimates of the $\bW$ and $\bV$ parameters.  Traditionally, the NN community uses  {\it backpropagation} to do this.  Backpropagation is based on applying the chain rule to calculating the gradient, which is straight-forward and useful due to the hierarchical/compositional nature of the model.  This is implemented in a two-pass algorithm that has the important feature of {\it locality}, in that each hidden unit passes and receives information only to and from units that share a connection.  This facilitates computation in a parallel computing environment, which is important for large datasets. 

Because the objective function consists of a sum over the training data, which can be quite large, computation of the gradient can be expensive. In addition, there may be redundant data in the training sample. One way to mitigate these issues is to consider minimizing the expected loss, which can easily be estimated by  averages of small random samples (i.e., minibatches) of the training sample.  This is the essence of {\it stochastic gradient descent (SGD)}, which is the dominant paradigm in modern neural computing \citep[e.g., see][]{goodfellow2016deep,aggarwal2018neural}.  Not only does it help with the big data, but SGD also helps keep the optimization from local minima.  Even these simple one layer NNs tend to overfit and it is important that they include some form of regularization.  For example, $L_2$ (ridge) penalties on the weights can be added to the objective function (known as ``weight decay'') or $L_1$ (lasso) penalties can be added, which is called ``weight elimination.''


\subsection{Deep Feedforward Networks (DNNs)}\label{sec:DNN}

Many problems that have big data, such as acoustic processing, image processing, and natural language processing, have very complex structure and have provided the motivation for the development of a new generation of deep learning algorithms. These are typically NNs with many hidden layers, with outputs from one layer becoming the input to the next.  We consider the number of units in each layer as the {\it width} and the number of layers the {\it depth} of the network.  Having both width and depth provides a very flexible learning environment, but brings with it many challenges.   DNNs utilize many of the technological innovations that underly many of the current applications of deep learning in large datasets \citep[e.g.,][]{hinton2012deep}.  Comprehensive overviews can be found in  \citet{goodfellow2016deep} and \citet{aggarwal2018neural}. 

A basic DNN can be represented as:
$$
{\bz}(\bx) = g_{o,\bV_L}(g_{\bW_L}(\cdots g_{\bW_1}(\bx))),
$$
where $g_{o,\bV_L}$ is an output function with weights $\bV_L$ and $g_{\bW_{\ell}}$ is a nonlinear activation function depending on parameters $\bW_{\ell}$ as in (\ref{eq:NNactivation}).   The hierarchical nature of a DNN is apparent in a simple example with two hidden layers and one output layer (with an identity output function): 
$$
\bz = \bV \by_2,
$$
$$
\by_2 = g(\bW_2 \by_1 + \bw_{0,2}),
$$
$$
\by_1 = g(\bW_1 \bx + \bw_{0,1}),
$$
where the dimension of the hidden vectors $\by_1$ and $\by_2$ may be different.  Training follows with backpropagation in an analogous way to the one-hidden layer model.  A significant challenge arises because there is typically a {\it huge} number of parameters in this model as the depth increases, which makes DNNs difficult to train.  In particular, in traditional applications with relatively small numbers of labeled responses there are several issues: e.g., (1) sensitivity to the number of hidden layers and number of hidden units; (2) sensitivity to other tuning parameters (one can use cross-validation if feasible); (3) extreme sensitivity to the initial values of the weights; (4) optimization is very slow on standard computation platforms; and (5) the fitted models have a propensity to overfit.

Modifications to the basic gradient-based optimization have allowed these models to be fit to large datasets.  One of the first ``breakthroughs'' was {\it generative pre-training}.  In essence, this is an attempt to get the parameters ``in the ball park'' before performing the backpropagation optimization.  The key idea behind {\it generative pre-training} is that one learns one layer at a time with the hidden units predicted at one level then serving as the input for training the next level.  This is {\it generative} in the sense that it starts at the bottom and builds one layer at a time -- ultimately {\it generating} a response.  The important thing here is that the associated estimates of the weights (which are approximations) just serve as starting values for the backpropagation algorithm.  The backpropagation algorithm then uses all of the information and fine tunes the estimates.  It is important to note that the generative pre-training does not use labeled responses, so it is unsupervised. This gives the parameters more freedom and prevents overfitting, but  the backpropagation algorithm uses the labeled responses to get the final estimates.  The primary generative models are {\it restricted Boltzman machines (RBMs)} and {\it autoencoders} \citep[e.g.,][]{goodfellow2016deep}.  Both of these approaches have the advantage that they have undirected connections (which guide the weights towards minima that improve generalization, e.g., see \citet{erhan2010does}), are easily stacked (so that the output of one can form the input for another), and are unsupervised.

In addition to the generative pre-training, other factors have proven important for the implementation of feedforward DNNs such as: (1) use of unlabeled data to train the model (this allows more flexibility); (2) use of {\it node dropout} for regularization (shrinkage), which helps dramatically with overfitting (essentially, each node has a probability of being in the model when being trained); (3) efficient computation (i.e., these models require a lot of computational power to fit --  distributed and parallel computing is essential, which has been made possible by graphical processing unit (GPU)-based parallel computing in recent years); and (4) rectified linear unit (ReLU) activation functions, $ReLU(x) = max(0,x)$, which can lead to faster training.  These models have perhaps shown the greatest success when they can also exploit the inherent multiscale nature of time and space, as with CNNs and RNNs.

\subsection{Convolutional Neural Networks (CNNs)}\label{sec:CNN}

\begin{figure}
\begin{center}
\includegraphics[width=2.5in,angle=-90]{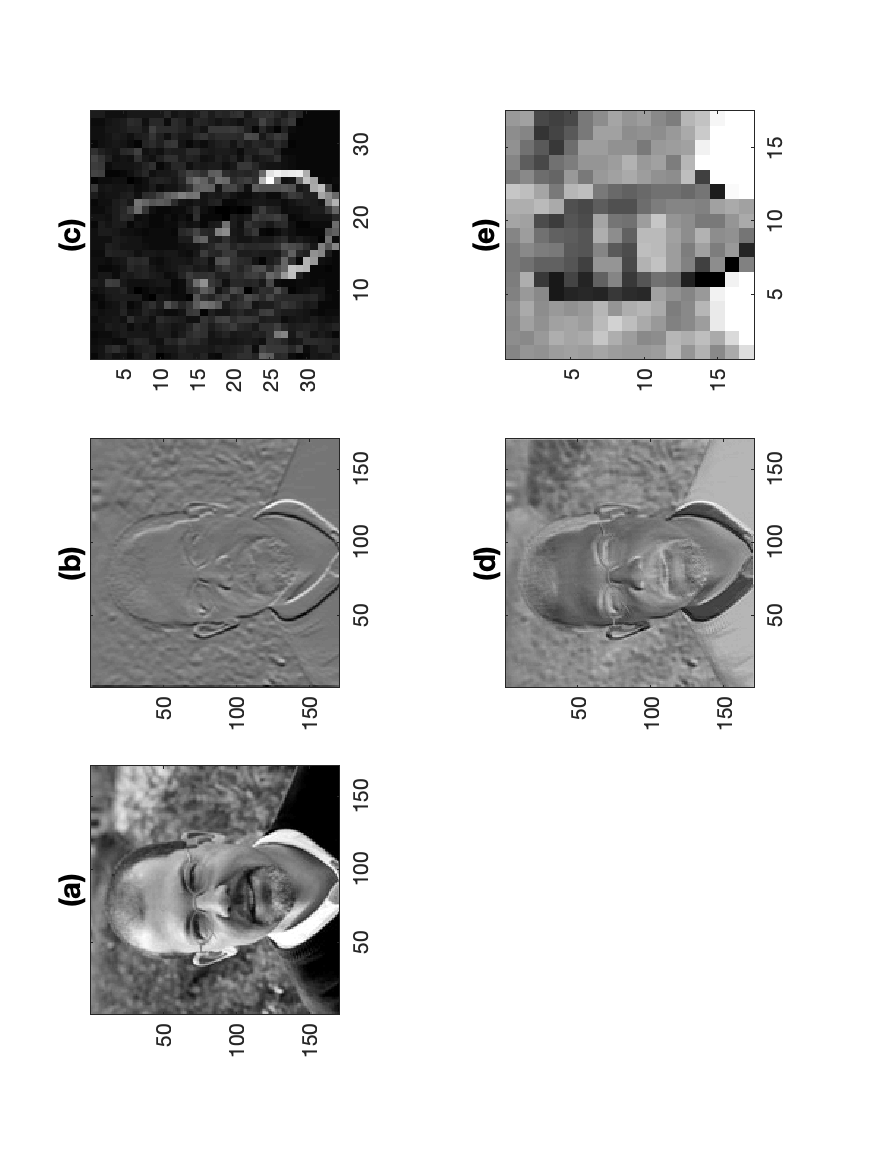} 
\caption{(a) Black and white image of the author; (b) convolution of image in (a) with a $3 \times 3$ Sobel edge detection (x-direction)  kernel; (c) $5 \times 5$ max pooling of the convolved image in (b);  (d) convolution of image in (a) with a $3 \times 3$ random ($unif(-0.1,0.1)$) kernel; (e) $10 \times 10$ median pooling of the convolved image in (d). }
\label{fig:convexample}
\end{center}
\end{figure}

One of the biggest success stories in deep learning has been CNNs, especially in the context of  image processing.  Recall the definition of a discrete convolution in two dimensions:
$$
k[x,y] \ast z[x,y] = \sum_{i=-\infty}^{\infty} \sum_{j=-\infty}^{\infty} k[i,j] gzx-i,y-j],
$$
where in practice, because there are a finite number of pixels in an image, the sums are finite.  We can think of $k[ \;]$ as a kernel weight function that is applied  to elements of the spatial image $z[ \; ]$.  Depending on the kernel weights, one can get different properties associated with the image after doing the convolution (see Figure \ref{fig:convexample}).  Note, in practice, color images have pixels represented by a combination of red, green, and blue (RGB) pixels, so images are best thought of as tensors.   One can easily modify the convolution function to operate on tensor-valued pixels.  

The CNN considers a convolution of the image with unknown weights that are learned; this is done multiple times for each level to get different ``feature maps.'' 
That is, rather than specify kernel functions, CNNs learn them in a way that there is one set of kernel weights for each convolution, so the weights are shared across the image (this leads to a significant dimension reduction in the number of parameters that must be learned).  This convolution step is then followed by a {\it pooling layer} (or, {\it subsampling} or {\it down sampling}).  The pooling layer considers a small rectangular block from the convolutional step and subsamples or aggregates it in some way to produce a single output. Perhaps the most common pooling simply takes the block maximum (known as ``max pooling'').  Pooling is beneficial because it helps make the CNN less sensitive to translations of the input. Importantly,  it also reduces the size of the next level image.  The right panels in Figure \ref{fig:convexample} illustrate pooling.

The general structure of a CNN has alternating convolution layers followed by pooling layers, with the last layer being fully connected (as in the DNN).  Typically, there are 
(1) multiple feature maps at the convolution stage created via multiple kernel weight matrices; (2) the convolved images go into a nonlinear activation function -- usually, a ReLU function; and (3) pooling can occur across multiple feature maps.  The critical stage of the CNN that requires estimation is the convolution step. Let $y_{i,j}^{\ell-1}$ correspond to the input to a convolutional step.  The convolution is then given by:
$$
y_{i,j}^{\ell} = g_p(g(\sum_a \sum_b k_{a,b}^{(\ell)} y_{i+a,j+b}^{\ell-1})),
$$
where $g_p(\cdot)$ is the pooling function, $g(\cdot)$ is a nonlinear activation (e.g.,ReLU), and $k_{ab}$ are the kernel weights that must be learned (estimate).  Note, the pooling layers are simple and are not learned.  As with DNNs, training the other components of the model is accomplished through a gradient descent back propagation algorithm, with the same enhancements described in Section \ref{sec:DNN}.

\subsection{Recurrent Neural Networks (RNNs)}

Recurrent Neural Networks (RNNs) were originally developed in the 1980s to process sequence data.  In recent years, they have been enhanced to be one of the most used and successful deep learning methods, particularly for language processing applications (e.g., speech recognition, text generation, machine translation, etc.).  These models are analogous to multivariate state-space models for dynamical systems, as one might see in time series, econometrics, or spatio-temporal statistics.  Consider a classical dynamical system:
$\by_t = {\cal M}(\by_{t-1};{\mbv \theta})$, where $\by_t$ represents the state of the system at time $t$.  This is considered  ``recurrent'' because the state at time $t$ refers back to the state at time $t-1$, etc.  We can rewrite this in a so-called ``unfolded'' form, $\by_t = {\cal M}({\cal M}({\cal M}(\by_{t-3};\bftheta);\bftheta);\bftheta \cdots)$. Note, the parameters ${\mbv \theta}$ are shared across all states.  The hidden states are then related to an output $\bz_t$ in an observation equation.  As with state-space models, the states in the RNN setting are likely to depend on external inputs, $\bx_t$. 

Thus, the most basic (``vanilla'') RNN is given by
$$
\bz_t = g_o(\bV \by_t)
$$
$$
\by_t = g(\bW \by_{t-1} + \bU \bx_t),
$$
where $g_o(\cdot)$ is an output function, $g(\cdot)$ is typically a hyperbolic tangent activation function, and $\bU$, $\bV$, and $\bW$ are weight matrices (which typically contain bias/offset terms as well).  As with other NNs, to estimate the parameters, one defines a loss function and would like to optimize by backpropagation with SGD.  However, there is a complication in the case of RNNs because the parameters are common across time, and so one must implement a {\it backpropagation through time (BPTT)} algorithm \cite[e.g., see the overview in][]{aggarwal2018neural}.   A serious challenge to implementing the BPTT optimization for this vanilla RNN is the so-called {\it vanishing gradient/exploding gradient} problem.  That is, the gradient can become increasingly smaller (typically) or larger as one moves through each time step (there are typically many time steps in an RNN implementation). 

There are a number of modifications to RNNs that have been specified to mitigate the vanishing/exploding gradient problem.  Perhaps the most common approach includes  {\it gates} that break up the temporal structure, allowing some hidden states in the past to be considered at certain time steps and others to be forgotten.  For example, the  {\it long short-term memory} (LSTM) RNN uses gates to create time paths that have gradients that do not vanish or explode \citep{hochreiter1997long}.  The basic LSTM structure is given as (note, $\circ$ is the Hadamard (element-wise) product):
$$\mbox{Output:}\;\;\; \bz_t = g_o(\bV \by_t)$$
$$ \mbox{Hidden State: \;\;\;} {\mbf y}_t = \tanh({\mbf c}_t) \circ {\mbf o}$$
$$ \mbox{Internal Memory: \;\;\;} {\mbf c}_t = {\mbf c}_{t-1} \circ {\mbf f} + {\mbf g} \circ {\mbf i}$$
 $$ \mbox{Candidate Hidden State: \;\;\;}  {\mbf g} = \tanh(\bU^g \bx_t + \bW^g \by_{t-1})$$
 $$ \mbox{Output Gate: \;\;\;}  {\mbf o} = g(\bU^o \bx_t + \bW^o \by_{t-1})$$
$$ \mbox{Forget Gate: \;\;\;}  {\mbf f} = g(\bU^f \bx_t + \bW^f \by_{t-1})$$
$$ \mbox{Input Gate:\;\;\; } {\mbf i} = g(\bU^i \bx_t + \bW^i \by_{t-1}),$$
 where typically $g(\cdot)$ is a sigmoid function.   The {\it input gate} selects hidden units that get input to time $t$, the {\it forget gate} selects the hidden states at the previous time to reset to 0 at time $t$, and the {\it output gate} selects the states that will be related to the response.   The memory units are crucial as they indicate when to remember or forget previous hidden states -- this memory feature is not only helpful in mitigating the vanishing/exploding gradient problem, but realistic for many processes in which events in the (distant) past can influence the presence irrespective of the intervening states.  A slightly simpler gated RNN that has gained recent popularity is the {\it gated recurrent unit} (GRU) RNN \citep{cho2014learning}. 
 
 In general practice, gated RNNs can be computationally intensive and often require parallelized implementations, and like the standard DNN and CNNs, require a lot of training data. 
 In the literature, the gated algorithms are considered more as ``black boxes'' given their complexity, which has the benefit of making them modular and connectable (see Section \ref{sec:DN-DSTMs}).
 
 \subsubsection{Echo State Networks (ESNs)}\label{eq:ESN}

An  alternative RNN that is easy to estimate and typically requires less computational resources and training data is the {\it echo state network (ESN)} \citep{lukovsevivcius2009reservoir}:
$$
\bz_t = g_o(\bV \by_t)
$$
$$
\by_t = g(\bW^* \by_{t-1} + \bU \bx_t).
$$
This looks like the basic RNN given above, but remarkably, the weight matrices $\bW$ and $\bU$ are sparse and chosen randomly in the ESN, so only the output matrix $\bV$ is learned (with regularization).  This use of random parameters in the nonlinear transformation is referred to generally as ``reservoir computing.''  One complication is that this approach requires a modification of the weights $\bW$ (given here by $\bW^*$ -- see below) to ensure the ``echo state property,'' which essentially states that the effects of the initial conditions diminish asymptotically with time.  Overall, the ESN provides an enormous reduction in parameters to be estimated and greatly simplifies the model so that $\by_t$ is simply a series of stochastic transformations of the inputs $\bx_t$ based on random weights, and the $\bV$ parameters in the output function $g_o(\cdot)$ can be trained as in basic statistical models (e.g., regression, logistic, softmax).  The ESN usually requires more hidden units than a traditional RNN (i.e., is wider) to compensate for not learning the weights, and so one has to apply regularization when estimating $\bV$.  We discuss ESN models in greater detail in the context of DSTMs in Section \ref{sec:linking} below.

\subsection{Deep Neural DSTMs (DN-DSTMs)}\label{sec:DN-DSTMs}


Although DNNs can be used with spatio-temporal data \citep{polson2017deep}, they are not always appropriate because they do not naturally accommodate dependence structures that occur in time and space. However,
given the modularity of CNNs and RNNs (i.e., they are easily ``stacked'' to make deeper models) it is no surprise that they can easily be combined in different ways to produce deep hybrid models for spatio-temporal data, such as video image processing and image captioning \cite[e.g.,][]{ keren2016convolutional,tong2018reservoir}.   For example, images in a video can be reduced by a CNN to find spatial features and the time evolution of these features can then be modeled with an RNN (usually an LSTM).  In some cases, this framework can also be used to relate images to captions or descriptions \citep{donahue2015long}.  That is, the CNN is used to encode the image and the RNN is used to decode relative to a sequence of words that describes the image.   The first case is clearly a spatio-temporal problem and the last is ``temporal" in the context that the output (a sequence of words) has a sequential structure.  In general, the ability of software packages to modularize the various machine learning components (such as CNNs and RNNs) allows developers to combine these layers in different ways.   Here, our interest is with spatial processes evolving through time (i.e., analogous to the first scenario).   Such approaches have been used in environmental science to produce nowcasts of precipitation \citep{xingjian2015convolutional}.

\begin{figure}[h]
\begin{center}
\includegraphics[width=2.5in]{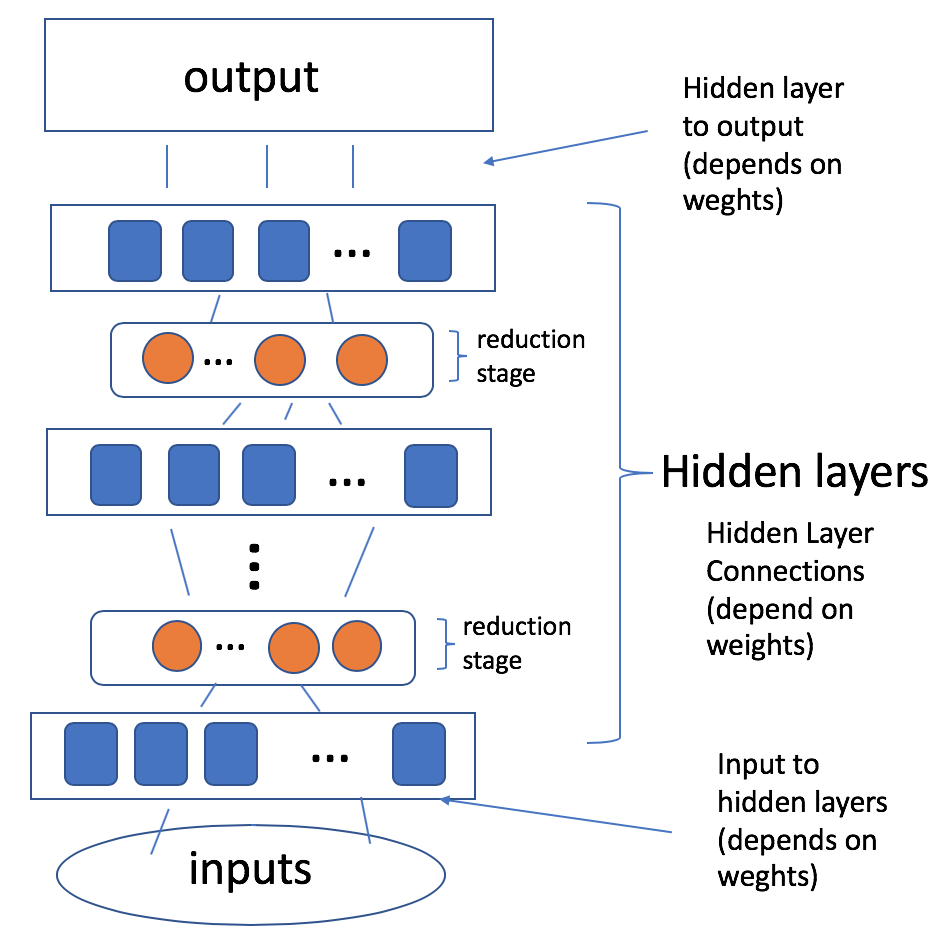} 
\caption{Schematic representation of a general deep neural dynamic spatio-temporal model (DN-DSTM). }
\label{fig:deepmodel}
\end{center}
\end{figure}

A general approach to the hybrid DN-DSTM considers a stacked RNN but with intermediate layers that reduce dimension.  This is shown schematically in Figure \ref{fig:deepmodel} and can be written generally as:
\begin{eqnarray}
 & \; & \text{\bf Output State:} \;\; \;\; {\mbf z}_t = g_o\Big(\by_{t,1},\widetilde{\by}_{t,2},\ldots,\widetilde{\by}_{t,L}; \bftheta_z \Big),\label{GenNN-DSTN}\\
  & \; & \text{ \bf Hidden Stage 1:}   \;\;   \;\;  {\by_{t,1}=g \Big(\by_{t-1,1}, \widetilde{\by}_{t,2}; \bftheta_{h1} \Big) } \nonumber ,\\
    & \; & \text{ \bf Reduction Stage 1:}    \;\;   \;\;  { \widetilde{\by}_{t,2} \equiv \mathcal{Q}(\by_{t,2};\bftheta_{r1}),} \nonumber \\
      & \; & \text{ \bf Hidden Stage 2:}   \;\;   \;\;  {\by_{t,2}=g \Big(\by_{t-1,2},\widetilde{\by}_{t,3}; \bftheta_{h2}  \Big) }, \nonumber \\
          & \; & \text{ \bf  Reduction Stage 2:}    \;\;   \;\; {  \widetilde{\by}_{t,3} \equiv \mathcal{Q}(\by_{t,3};\bftheta_{r2}), } \nonumber \\
                           & \; &   \;\;   \;\;   \;\;  \;\;   \;\; { \mbox{ \Large {\mbv \vdots}} }  \;\;   \;\;   \;\;   \;\;   \;\;      \;\;   \;\;   \;\;   \;\;   \;\;   \;\;   \;\;     { \mbox{ \Large {\mbv \vdots}} } \nonumber  \\
    & \; & \text{ \bf Hidden Stage L:}  \;\;   \;\;  \;\;   \;\;  \by_{t,L}=g \Big(\by_{t-1,L}, \widetilde{\bx}_t;\bftheta_{hL}  \Big) \nonumber, \\
      & \; & \text{ \bf Input Stage:}  \;\;   \;\;  \;\;   \;\;  { \widetilde{\bx}_{t}= g_I \Big(\bx_t;\bftheta_{I}  \Big)} \nonumber, 
\end{eqnarray}
where  $g_o(\cdot)$ is an output function (e.g., identity for regression, softmax for classification, etc.), $g_I(\cdot)$ is an input function that potentially augments and/or transforms the input vector $\bx_t$, $g( \cdot)$ is some type of RNN structure (e.g., LSTM, GRU, ESN),  and $\mathcal{Q}( \cdot)$ is a dimension reduction function such as a CNN or something simpler such as a principal component decomposition or some other stochastic dimension reduction approach (e.g.,  random projection, \citet{bingham2001random}).  The potential parameters (weights) in each function are given by the $\bftheta$s.  In this framework, the components at each reduction stage, $\tilde{\by}_{t,\ell}$, can influence the output, in addition to the non-reduced hidden units from Stage 1.  One could also have the non-reduced hidden units from the deeper hidden stages influence the output directly as well, but this increases the number of parameters that must be learned and is typically not necessary.  Finally, note that this model might be written more concisely as a telescoping functional transformation of the input:
\begin{equation}
{\mbf z}_t = g_o(g(\mathcal{Q}(g( \cdots \mathcal{Q}(g(g_I(\bx_t))))));{\mbv \Theta}),
\label{eq:telescopSTNN}
\end{equation}
where ${\mbv \Theta}$ represents all of the various parameters (weights) in the functions.

The advantage of such an approach is that it naturally accommodates multiple spatial and temporal scales of variability.  Note, $g_I(\cdot)$ acts as an encoder that transforms the inputs. For example, $g_I(\cdot)$ might be a CNN or it might be some other type of dimension reduction procedure (e.g., principal components, Laplacian eigenmaps, kernel convolutions, etc.).  Then the $\mathcal{Q}$ functions extract important dependent features in the hidden units (that may be spatially referenced depending on the choices of $g_I(\cdot)$, $g(\cdot)$ and $\mathcal{Q}$.   The various RNN levels then act to find temporal dependencies, typically at different scales in time \citep[e.g.,][]{graves2013speech,hermans2013training}.  Note that one can leave out various levels; e.g., we might leave out a $\mathcal{Q}$ stage and form a stacked RNN without the intervening reduction stage (and, vice versa).  Typically, such a model would be implemented via back propagation and SGD, depending on the choice for different model stages.  


\subsection{Connections between DH-DSTMs and DN-DSTMs}\label{sec:connections}

The natural question is how do the DH-DSTMs presented in Section \ref{sec:H-DSTMs} compare to the DN-DSTMs presented in Section \ref{sec:DN-DSTMs}?  
The two paradigms do have much in common in that they are both trying to do the same thing in the context of modeling complex spatio-temporal dependence.  That is, both are dealing with the fact that there are multiple scales of spatio-temporal variability that interact to describe process evolution and are building that complex dependence in some sense by ``marginalizing'' common components.    Specifically, both model frameworks: (a) consist of multiple connected telescoping levels; (b) include dimension reduction stages; (c) typically do not model second order dependence (note, GP networks and restricted Boltzmann machines are an exception); (d) can handle multiple inputs (predictors) and different output types; (e) have a very large number of parameters to estimate; (f) require a lot of training data; (g) require prior information (or, pre-training, heuristics, etc.); (h) require regularization; (i) are expensive to compute and require efficient algorithmic implementations. 

The aforementioned points suggest that one of the main challenges for both the DH-DSTM and DN-DSTM frameworks is related to implementation and computation.  That is, in the DH-DSTM framework, one must make many decisions concerning the types of dependence structure, whether to put structure in the covariance or the mean, the amount of mechanistic information to include, and the prior distributions, just to name a few.  In addition, in these complex modeling situations, one typically must program the DH-DSTM from scratch in some relatively efficient language as the automated packages that perform Bayesian computation are often not flexible enough to accommodate DH-DSTMs, or are too inefficient (i.e., their strength in providing general solutions can be a limitation for certain specific dependence structures).  Similarly, the DN-DSTM models can also have a very large number of tuning parameters and model choices (e.g., choice of $g( \cdot)$, $\mathcal{Q}$, the number of layers, the number of hidden units per layer, the type of regularization, pretraining, etc.).  Although the aforementioned references contain suggestions for some cases, there is no universal advice for these decisions -- it is very much an experience and trial--and-error endeavor.  However, unlike with DH-DSTMs, there are standard software environments such as Tensor Flow, Theano, Caffe, pyTorch (and many more!) that are quite flexible and, in some sense, modular, which has increased their utility in production environments.

There are a number of other structural differences between the modeling paradigms. First, the DH-DSTM framework is based on stochastic models that include distributional error terms within a valid probability construct (i.e., the joint distribution of all random components can be written as a series of conditional models). In contrast, the DN-DSTM framework is deterministic with no error terms (note the caveat that when one uses reservoir methods (e.g., ESNs for $g( \cdot)$), then (\ref{eq:telescopSTNN}) is a stochastic transformation but not a formal stochastic model).   One consequence to the lack of a probabilistic structure for the DN-DSTM is that there is no clear mechanism to produce model-based estimates of uncertainty in the prediction or classification that results from the DN-DSTM.  Second, one is limited in performing inference on the parameters -- although, it should be noted that this would seldom be of interest in this type of model as the parameters are typically not identifiable, highly dependent, and non-interpretable. 

In addition, it is still an open problem on how to generally include known relationships (e.g., such as suggested by a mechanistic model) in the deep NN framework \citep[although, see][for recent work in this area]{karpatne2017theory}.  That said, the DN-DSTM framework does have some important advantages in that it is easy to manipulate and implement different model structures (e.g., stacking different model components) in the backpropagation estimation paradigm implemented in many of the existing software packages.  Finally, in the context of spatio-temporal dynamics, it should be noted that the RNN structure can naturally accommodate non-Markovian dynamics (e.g., memory of distant past events).  This last point is potentially important to environmental, ecological, and agricultural applications and has not been a concentrated focus in statistical implementations of spatio-temporal models.  

\section{Combining the DH-DSTM and DN-DSTM Frameworks}\label{sec:linking}

A natural approach to combine the DH-DSTM and DN-DSTM frameworks would be to allow the parameters in the DN-DSTM to be random, perhaps add some error terms, and then implement via a Bayesian paradigm.  Although Bayesian implementations of neural nets have been considered at least since the 1990s \citep{mackay1992practical,neal1996bayesian}, it is exceedingly challenging to implement deep neural models from a fully Bayesian perspective due to the extremely large number of dependent and non-identifiable parameters \citep[see the overview in][]{polson2017deepB}.   Such models can be implemented in some contexts \citep[e.g.][]{chatzis2015sparse,chien2016bayesian,gan2016scalable,mcdermott2017bayesian} but are quite sensitive to particular data sets and are typically computationally prohibitive.  More recently, approximate Bayesian methods such as variational Bayes \citep{tran2018bayesian}, and scalable Bayesian methods \citep{snoek2015scalable} have been used successfully in deep models.  In the context of DN-DSTMs this is still an active area of research.

Alternatively, two relatively simple approaches have recently been used to blend the DN-DSTM and DH-DSTM paradigms.  These do so in a way that also mitigates the challenges associated with implementing DH-DSTMs.  That is, DH-DSTMs typically suffer from a curse of dimensionality in parameter space and require a large amount of data and fairly specialized computational algorithms and, thus, are fairly inefficient to develop and implement.  The hybrid approaches mitigate these issues but still provide a flexible and effective approach to model complex spatio-temporal processes in a manner that accounts for uncertainty quantification. 

\subsection{An Ensemble Approach}

\citet{mcdermott2017ensemble} made several modifications to the standard ESN model to account for a simple approach to 
uncertainty quantification in a spatio-temporal nonlinear forecasting setting.  They considered a {\it quadratic ESN model}.  That is,
for $t=1,\ldots,T$, let:
\begin{eqnarray}
\mbox{Response: }  & \; & \bz_t = \bV_1 \by_t + \bV_2 \by^2_t + {\mbv \epsilon}_t, \quad \textrm{for} \quad {\mbv \epsilon}_t \; \sim \; Gau({\mbf 0},\sigma^2_\epsilon \bI); \label{eq:QESNresp}\\
\mbox{Hidden State: }  & \; &  \by_t = g\left(\frac{\nu}{|\lambda_w|}{\bW} {\by}_{t-1} + {\bU}\tilde{\bx}_t\right); \label{eq:QESNhidden} \\
\mbox{Parameters: } & \; & {\bW} = [w_{i,\ell}]_{i,\ell}: w_{i,\ell} = \gamma^w_{i,\ell} \; Unif(-a_w,a_w) + (1 - \gamma^w_{i,\ell}) \; \delta_0, \label{eq:QESNw} \\
& \; & {\bU} = [u_{i,j}]_{i,j}: u_{i,j} = \gamma^u_{i,j} \; Unif(-a_u,a_u) + (1 - \gamma^u_{i,j}) \; \delta_0, \label{eq:QESNu}  \\
& \; & \gamma_{i,\ell}^w \; \sim \; Bern(\pi_w), \label{eq:piw} \\
& \; & \gamma_{i,j}^u \; \sim \; Bern(\pi_u) \label{eq:piu},
\end{eqnarray}
where $g(\cdot)$ is an activation function (usually a hyperbolic tangent function),  $\lambda_w$ is the ``spectral radius'' (the largest eigenvalue of $\bW$), and $\nu$ is a scaling parameter taking values between $[0,1]$ that helps control the amount of memory in the system, $\bW$, $\bU$, $\bV_1$, and $\bV_2$ are weight matrices,  $\delta_o$ is a Dirac function, $\gamma^w_{i,\ell}$, $\gamma^u_{i,\ell}$ denote indicator variables, and $\pi_w$, $\pi_u$ represent the probability of a parameter in the weight matrices being 0.  Note, dividing by the spectral radius in \ref{eq:QESNhidden} ensures the echo state property mentioned previously, and $\nu$ controls the memory. The only parameters that are estimated in this model are those in $\bV_1$ and $\bV_2$, and $\sigma^2_\epsilon$ from Equation (\ref{eq:QESNresp}), for which we use a ridge penalty hyperparameter, $r_v$.  Again, it is important to note that $\bW$ and $\bU$ are not estimated, but simply drawn from (\ref{eq:QESNw}) and (\ref{eq:QESNu}), respectively. The hyperparameters $\pi_w$, $\pi_u$, $a_w$, $a_u$, $\nu$, and $r_v$ are specified as discussed below. 

The modifications of the ESN that make it useful as a DSTM are the inclusion of the explicit error term, ${{\mbv \epsilon}_t}$, the quadratic term $\bV_2 \by^2_t$ and, most importantly, vector {\it embeddings} of the inputs:
\begin{equation}
\tilde{\mbf x}_t = [\bx'_t,\bx'_{t-\tau}, \bx'_{t- 2 \tau},\ldots,\bx'_{t - m \tau}]'. \nonumber
\end{equation}
 An embedding includes lagged values of the input predictor and is important due to Takens' theory \citep{takens1981detecting} in dynamical systems that states that one can represent a state space of high dimension by a sufficiently large number of lagged values of a portion of the state space.   Note that the results are not very sensitive to $\{\pi_w, \pi_u, a_w,a_u\}$  and they are usually fixed at small values, but the results can be sensitive to $\{n_h, \nu, r_v \}$, so they are chosen by cross-validation.

\begin{figure}
\begin{center}
\includegraphics[width=0.8\textwidth,keepaspectratio=true]{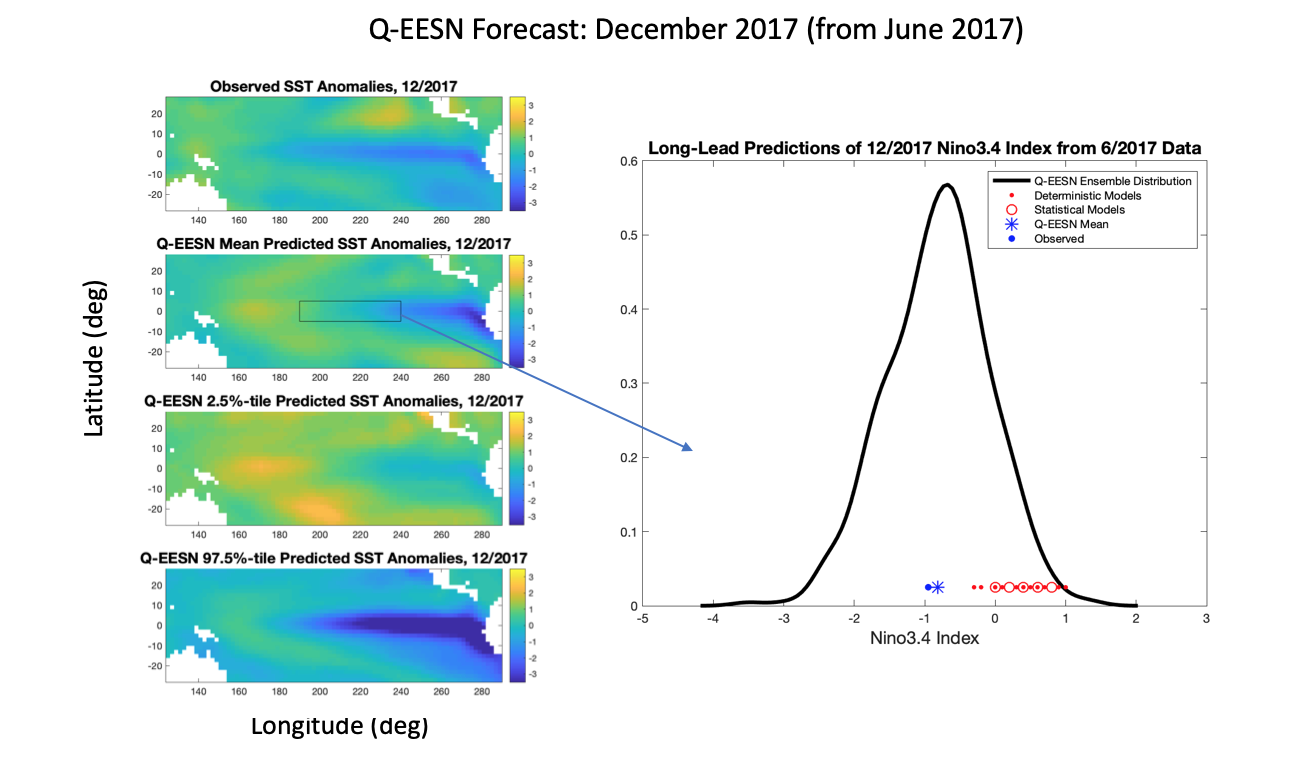}
\caption{Left Panels: Long-lead forecast summary maps for Pacific SST for 6-month forecasts valid in December 2017 given observations through June 2017.  The top row shows the observed SST anomalies  (deviations from climatological average).  The forecast mean from the Q-EESN model is shown in the second panel, and the bottom two panels show the lower and upper quantiles for a 95\% prediction interval calculated in each grid cell.  The figure on the right shows the Q-EESN predictive distribution for the so-called Nino3.4 index, which is based on an average in the region denoted by the box in the second panel on the left side.  The blue star shows the Q-EESN forecast mean and the observed index value  is denoted by the filled blue circle. The solid and open red circles correspond to forecasts based on the same starting and verification period for  deterministic and stochastic models presented by IRI/CPC (see the footnote in the text for the website). }
\label{fig:longlead_sst}
\end{center}
\end{figure}

\citet{mcdermott2017ensemble} consider a simple ensemble forecast approach (analogous to a parametric bootstrap; \cite{sheng2013prediction},  in which multiple samples from the reservoir matrices $\bW$ and $\bU$ are drawn and the model is refit for each parameter set.  This gives a distribution of the output predictions and allows the quantification of uncertainty in the predictions.  They present an example in which this quadratic ensemble ESN (Q-EESN) model is used to generate long-lead (6 month) forecasts of tropical Pacific SST (i.e., El Ni\~no and La Ni\~na events).  The model performed very well.  For example, Figure \ref{fig:longlead_sst} shows the prediction and prediction uncertainty for a forecast of SST in December 2017 given data through June 2017 (which exhibited a  La Ni\~na event).  Note, however, that the dynamical and statistical forecasts presented for this same period by the US National Oceanic and Atmospheric Agency's Climate Prediction Center (CPC) and International Research Institute  (IRI) for Climate and Society at Columbia University\footnote{\url{https://iri.columbia.edu/our-expertise/climate/forecasts/enso/2017-July-quick-look/?enso_tab=enso-sst_table}} did not suggest a La Ni\~na would develop (their probability forecast was around 15\% for a La Ni\~na for this period).   The reasons for the success  of the Q-EESN approach here are likely related to the fact that the ESN is a dynamic model that incorporates nonlinear interactions, but also that it augments the input space to perform a regression \citep{gallicchio2011architectural}.  That is, the dimension of $\by_t$ is typically larger than $\tilde{\bx}_t$ (i.e., a dimension expansion of the potential predictors).  In addition, the small, sparse, random weights limit overfitting and regularize the regression.  Finally, the embedded inputs in the Q-EESN implementation allow for additional nonlinearity, and the ensemble bootstrap approach with relatively few hidden units provides a  ``committee of weak learners.''  It is important to note that this approach takes just seconds to implement on a laptop computer compared to hours for traditional  DH-DSTM approaches.



\subsection{A Deep Basis Function Approach}
The Q-EESN model has no mechanism to link hidden layers, which are important for processes that occur on multiple time scales. There have been deep ESN models implemented in the ML literature \citep[e.g.,][]{jaeger2007discovering,triefenbach2013acoustic,antonelo2017echo, ma2017deep,gallicchio2018design}, but these approaches generally do not  accommodate uncertainty quantification and are not designed for spatio-temporal systems.   However, one could extend these deep ESN models to accommodate spatio-temporal processes as in (\ref{GenNN-DSTN}).  For example, \citet{mcdermott2018deep} did this within an ensemble parametric bootstrap context to account for multiple time scales and uncertainty in predictions.  They also consider an implementation where (\ref{GenNN-DSTN}) is used to generate basis functions that are a stochastic transformation of the inputs.  This is especially useful in a spatio-temporal regression context, i.e., when one seeks to predict one spatio-temporal process based on another.   Specifically, consider the model:
\begin{eqnarray*}
 & \; & \text{ Data Stage:}   \;\;   \;\;  { \bz_t \sim Gau({\mbv \Phi} {\mbv \alpha}_t, \bC_z)}  \\ 
 & \; & \text{ Output Stage:} \;\;   \;\; { {\mbv \alpha}_t = \sum\limits_{j=1}^{n_{res}} \big[  {\mbv \beta}_1^{(j)}\by_{t,1}^{(j)} +  \sum\limits_{\ell=2}^L {\mbv \beta}_\ell^{(j)}  \widetilde{\by}_{t,\ell}^{(j)} \big] +{\mbv \eta}_t,} \;\; {{\mbv  \eta}_t \; \sim \; \text{Gau}({\mbf 0},\sigma^2_ \eta \bI)}, \label{eq:BD-EESNoutput}\\
 &\; &  \text{ Priors:}   \;\;   \;\;  { \beta^{(j)}_{\ell,b} \mid \gamma_{\ell}^{\beta_\ell} \sim \gamma_{\ell}^{\beta_\ell} \;  \text{Gau}(0,\sigma^2_{\beta_\ell,0}) +  (1- \gamma_{\ell}^{\beta_\ell} ) \; \text{Gau}(0,\sigma^2_{\beta_\ell,1}),}  \nonumber \\
 &\;& \;\; \;\; \;\;\;\;\;\;\;\;\;\;\;\;\;\;\;\; { \gamma_{\ell}^{\beta_\ell}\sim \text{Bernoulli}(\pi_{\beta_\ell}), }  \nonumber \\
 &\;& \;\; \;\; \;\;\;\;\;\;\;\;\;\;\;\;\;\;\;\;  { \sigma^2_\eta \sim \text{IG}(\alpha_\eta,\beta_\eta),} \nonumber
\end{eqnarray*}
where $\;\; {\by_{t,1}^{(j)}}$, ${ \widetilde{\by}_{t,\ell}^{(j)}}$ are a function of $\tilde{\bf x}_{t-\tau}$ as given in (\ref{GenNN-DSTN}), and ${\mbv \beta}_\ell^{(j)}$ are the associated regression coefficients for the $j$th ensemble and $\ell$th level.  Importantly, the $y$s are generated ``offline'' from an ensemble deep ESN with principal component reduction stages for $\cal{Q}$.  In addition, 
$$\{\pi_{w_1},\ldots,\pi_{w_L}, \pi_{u_1},\ldots,\pi_{u_L},a_{w_1},\ldots,a_{w_L}, a_{u_1},\ldots,a_{u_L}\}$$ 
are fixed at small values, and the number of hidden units for all layers except the first are  fixed since all of these layers go through the dimension reduction function $\cal{Q}$. Finally, 
$$\{\nu_1,\ldots,\nu_L, n_{\tilde{h},2},\ldots, n_{\tilde{h},L},n_{h,1}, r_\nu,m\}$$
are selected by a genetic algorithm.  The parametric bootstrap approach generates $j=1,\ldots,n_{res}$ ensembles of these deep ESNs by sampling different weight matrices as with the Q-EESN model (\ref{eq:QESNu}) and (\ref{eq:QESNw}) above.

As an example, \citet{mcdermott2018deep} consider  6 month long-lead forecasts of soil moisture over the US corn belt given Pacific SST.  Figure \ref{fig:longlead_sst} shows the out of sample forecast for May 2014 given SSTs from November 2017 based on a 3-level deep ensemble ESN model.  They show that this model performed the best compared to a variety of models in terms of a continuous ranked probability score and second best in terms of mean squared prediction error (the 2-level version of this model performed slightly better with this metric).
\begin{figure}
\begin{center}
\includegraphics[width=0.9\textwidth,keepaspectratio=true]{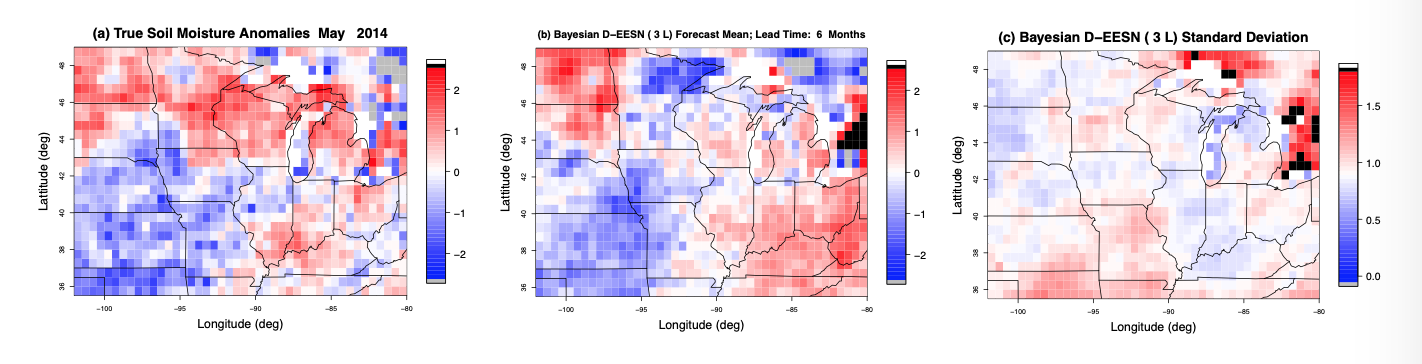}
\caption{Posterior summaries for the soil moisture long-lead prediction  in May 2014 using the 3-layer Bayesian deep ensemble echo state network model. (a) Observed soil moisture values for each spatial grid location. (b) Posterior predictive mean values for each grid location. (c) Posterior predictive standard deviations for each grid location.  Each plot has been standardized by their respective means and standard deviations to aid in visualization and extreme outliers have been removed for the sake of visualization (indicated by black grid squares). See  \citet{mcdermott2018deep} for details.}
\label{fig:longlead_sst}
\end{center}
\end{figure}

This approach is essentially a high-dimensional regression problem in which one generates a collection of basis functions by stochastic transformation of the inputs through the deep ESN model.  Multiple such transformations  are considered as potential predictors to give the approach flexibility and reproducibility.  The large number of predictors are controlled by SSVS regularization.  Note that the inputs (predictors) in this model are stochastically  {\it and dynamically} transformed. Thus,  the spatio-temporal regression model is not itself dynamic but, importantly, the transformations are dynamic through the ESN structure.  These multiple levels of  transformation allow for different time and spatial scales in the predictor variables to affect the response.  Importantly, by including the dynamics in the transformation (offline), this framework is very easy to implement through regularized regression methods and it is relatively efficient (compared to deep parametric statistical models and deep ML models) due to the reservoir approach in the ESN and simple regularization.   The data model here can easily accommodate other data types such as with deep Bayesian implementations of generalized linear mixed models  \citep[e.g.,][]{tran2018bayesian}.

\section{Discussion}\label{sec:discussion}

One of the fundamental principles of DH-DSTMs is that to model complex processes across multiple time and spatial scales, one benefits from considering a sequence of linked probability models.  In particular, because it is very difficult to specify the dependence structure for complex (e.g., nonlinear) spatio-temporal processes, one places modeling effort into the conditional mean and takes advantage of building dependence through marginalization.  Similarly, the deep neural models in ML that have become so popular in the last decade for image and language processing (e.g., DNNs, CNNs, RNNs) are also based on a sequence of linked models (typically, not stochastic models), with the outputs from one level becoming the inputs for the next.  The spatio-temporal version of these models, DN-DSTMs, typically combine CNNs and RNNs and also seek to build complexity by learning which scales of spatial and/or temporal variability are important for predicting responses.  These modeling frameworks have many practical issues in common, including the need for large training data sets, dimension reduction, regularization, and efficient computation.  Recent approaches to mitigate some of these issues, e.g., to apply the models when one does not have a huge amount of training data, have benefited from considering reservoir computing in the context of ESNs.  In spatio-temporal problems, these models have been placed in a statistical context through the use of parametric bootstrap and basis function transformation approaches.  These can be implemented at a fraction of the cost of traditional DH-DSTMs but still retain a probability formulation to allow uncertainty quantification and benefit from the flexibility of DN-DSTM's ability to flexibly model multiple time and spatial scales.

We have only scratched the surface in terms of blending the DH-DSTMs and DN-DSTMs for environmental, ecological and environmental statistics.  One important challenge is to be able to include mechanistic information efficiently in this blended framework.  Traditionally, it has been challenging to include such information in DN-DSTMs due to the conflict between mechanistic formulations and flexible learning formulations, and because of the challenge in training such models via gradient-based optimization.  In addition, there are potential advancements that can be obtained by including ideas from {\it deep reinforcement learning} \citep[e.g., see the overview in][]{aggarwal2018neural}.  Such methods train models in ways that they are rewarded for good decisions and penalized for poor decisions. This is the technology that was used for {\it AlphaGo} \citep{silver2016mastering} and later game-playing algorithms \citep{silver2018general}.  Useful connections to DH-DSTMs in environmental statistics are likely, given the long history of using reinforcement learning in control engineering.  In addition, it is likely that the hybridization of DH-DSTMs and DN-DSTMs can benefit from the recent advances in {\it generative adversarial networks} \citep{goodfellow2014generative}.  This approach trains models in a way  that benefits from two NNs competing against each other.  In particular, one network generates potential solutions and the other network evaluates or discriminates these solutions.   Indeed, the literature in deep neural modeling is advancing very rapidly, and it is exciting to see which of these methods and approaches can be included in more traditional probabilistic DSTM frameworks.

\section*{Acknowledgments}
This work was partially supported by the US National Science Foundation (NSF) and the US Census Bureau under NSF grant SES-1132031, funded through the NSF-Census Research Network (NCRN) program, and NSF award DMS-1811745.  The author would like to thank Brian Reich for encouraging the writing of this paper, Patrick McDermott for helpful discussions, and Nathan Wikle for providing helpful comments on an early draft.


\bibliography{reference}

\begin{thebibliography}{53}
\newcommand{\enquote}[1]{``#1''}
\expandafter\ifx\csname natexlab\endcsname\relax\def\natexlab#1{#1}\fi

\bibitem[{Aggarwal(2018)}]{aggarwal2018neural}
Aggarwal, C.~C. (2018), \textit{Neural networks and deep learning}, Springer.

\bibitem[{Antonelo et~al.(2017)Antonelo, Camponogara, and
  Foss}]{antonelo2017echo}
Antonelo, E.~A., Camponogara, E., and Foss, B. (2017), \enquote{Echo State
  Networks for data-driven downhole pressure estimation in gas-lift oil wells,}
  \textit{Neural Networks}, 85, 106--117.

\bibitem[{Berliner(1996)}]{berliner1996hierarchical}
Berliner, L.~M. (1996), \enquote{Hierarchical {B}ayesian time series models,}
  in \textit{Maximum Entropy and Bayesian Methods}, eds. Hanson, K.~M. and
  Silver, R.~N., Dordecht: Kluwer, Fundamental Theories of Physics, 79, pp.
  15--22.

\bibitem[{Bingham and Mannila(2001)}]{bingham2001random}
Bingham, E. and Mannila, H. (2001), \enquote{Random projection in
  dimensionality reduction: applications to image and text data,} in
  \textit{Proceedings of the seventh ACM SIGKDD international conference on
  Knowledge discovery and data mining}, ACM, pp. 245--250.

\bibitem[{Chatzis(2015)}]{chatzis2015sparse}
Chatzis, S.~P. (2015), \enquote{Sparse Bayesian Recurrent Neural Networks,} in
  \textit{Joint European Conference on Machine Learning and Knowledge Discovery
  in Databases}, Springer, pp. 359--372.

\bibitem[{Chien and Ku(2016)}]{chien2016bayesian}
Chien, J.-T. and Ku, Y.-C. (2016), \enquote{Bayesian recurrent neural network
  for language modeling,} \textit{IEEE transactions on neural networks and
  learning systems}, 27, 361--374.

\bibitem[{Cho et~al.(2014)Cho, Van~Merri{\"e}nboer, Gulcehre, Bahdanau,
  Bougares, Schwenk, and Bengio}]{cho2014learning}
Cho, K., Van~Merri{\"e}nboer, B., Gulcehre, C., Bahdanau, D., Bougares, F.,
  Schwenk, H., and Bengio, Y. (2014), \enquote{Learning phrase representations
  using RNN encoder-decoder for statistical machine translation,} \textit{arXiv
  preprint arXiv:1406.1078}.

\bibitem[{Cressie and Wikle(2011)}]{cressie2011statistics}
Cressie, N. and Wikle, C.~K. (2011), \textit{Statistics for Spatio-Temporal
  Data}, Hoboken, NJ: John Wiley \& Sons.

\bibitem[{Donahue et~al.(2015)Donahue, Anne~Hendricks, Guadarrama, Rohrbach,
  Venugopalan, Saenko, and Darrell}]{donahue2015long}
Donahue, J., Anne~Hendricks, L., Guadarrama, S., Rohrbach, M., Venugopalan, S.,
  Saenko, K., and Darrell, T. (2015), \enquote{Long-term recurrent
  convolutional networks for visual recognition and description,} in
  \textit{Proceedings of the IEEE conference on computer vision and pattern
  recognition}, pp. 2625--2634.

\bibitem[{Erhan et~al.(2010)Erhan, Bengio, Courville, Manzagol, Vincent, and
  Bengio}]{erhan2010does}
Erhan, D., Bengio, Y., Courville, A., Manzagol, P.-A., Vincent, P., and Bengio,
  S. (2010), \enquote{Why does unsupervised pre-training help deep learning?}
  \textit{Journal of Machine Learning Research}, 11, 625--660.

\bibitem[{Fan and Lv(2010)}]{fan2010selective}
Fan, J. and Lv, J. (2010), \enquote{A selective overview of variable selection
  in high dimensional feature space,} \textit{Statistica Sinica}, 20, 101.

\bibitem[{Gallicchio and Micheli(2011)}]{gallicchio2011architectural}
Gallicchio, C. and Micheli, A. (2011), \enquote{Architectural and markovian
  factors of echo state networks,} \textit{Neural Networks}, 24, 440--456.

\bibitem[{Gallicchio et~al.(2018)Gallicchio, Micheli, and
  Pedrelli}]{gallicchio2018design}
Gallicchio, C., Micheli, A., and Pedrelli, L. (2018), \enquote{Design of deep
  echo state networks,} \textit{Neural Networks}, 108, 33--47.

\bibitem[{Gan et~al.(2016)Gan, Li, Chen, Pu, Su, and Carin}]{gan2016scalable}
Gan, Z., Li, C., Chen, C., Pu, Y., Su, Q., and Carin, L. (2016),
  \enquote{Scalable Bayesian Learning of Recurrent Neural Networks for Language
  Modeling,} \textit{arXiv preprint arXiv:1611.08034}.

\bibitem[{Gelman and Hill(2006)}]{gelman2006data}
Gelman, A. and Hill, J. (2006), \textit{Data analysis using regression and
  multilevel/hierarchical models}, Cambridge university press.

\bibitem[{Gelman et~al.(2013)Gelman, Stern, Carlin, Dunson, Vehtari, and
  Rubin}]{gelman2013bayesian}
Gelman, A., Stern, H.~S., Carlin, J.~B., Dunson, D.~B., Vehtari, A., and Rubin,
  D.~B. (2013), \textit{Bayesian data analysis, third edition}, Chapman and
  Hall/CRC.

\bibitem[{Goodfellow et~al.(2016)Goodfellow, Bengio, Courville, and
  Bengio}]{goodfellow2016deep}
Goodfellow, I., Bengio, Y., Courville, A., and Bengio, Y. (2016), \textit{Deep
  learning}, vol.~1, MIT press Cambridge.

\bibitem[{Goodfellow et~al.(2014)Goodfellow, Pouget-Abadie, Mirza, Xu,
  Warde-Farley, Ozair, Courville, and Bengio}]{goodfellow2014generative}
Goodfellow, I., Pouget-Abadie, J., Mirza, M., Xu, B., Warde-Farley, D., Ozair,
  S., Courville, A., and Bengio, Y. (2014), \enquote{Generative adversarial
  nets,} in \textit{Advances in neural information processing systems}, pp.
  2672--2680.

\bibitem[{Graves et~al.(2013)Graves, Mohamed, and Hinton}]{graves2013speech}
Graves, A., Mohamed, A.-r., and Hinton, G. (2013), \enquote{Speech recognition
  with deep recurrent neural networks,} in \textit{Acoustics, speech and signal
  processing (icassp), 2013 ieee international conference on}, IEEE, pp.
  6645--6649.

\bibitem[{Heaton et~al.(2018)Heaton, Datta, Finley, Furrer, Guinness,
  Guhaniyogi, Gerber, Gramacy, Hammerling, Katzfuss, et~al.}]{heaton2018case}
Heaton, M.~J., Datta, A., Finley, A.~O., Furrer, R., Guinness, J., Guhaniyogi,
  R., Gerber, F., Gramacy, R.~B., Hammerling, D., Katzfuss, M., et~al. (2018),
  \enquote{A case study competition among methods for analyzing large spatial
  data,} \textit{Journal of Agricultural, Biological and Environmental
  Statistics}, 1--28.

\bibitem[{Hermans and Schrauwen(2013)}]{hermans2013training}
Hermans, M. and Schrauwen, B. (2013), \enquote{Training and analysing deep
  recurrent neural networks,} in \textit{Advances in neural information
  processing systems}, pp. 190--198.

\bibitem[{Hinton et~al.(2012)Hinton, Deng, Yu, Dahl, Mohamed, Jaitly, Senior,
  Vanhoucke, Nguyen, Sainath, et~al.}]{hinton2012deep}
Hinton, G., Deng, L., Yu, D., Dahl, G.~E., Mohamed, A.-r., Jaitly, N., Senior,
  A., Vanhoucke, V., Nguyen, P., Sainath, T.~N., et~al. (2012), \enquote{Deep
  neural networks for acoustic modeling in speech recognition: The shared views
  of four research groups,} \textit{IEEE Signal processing magazine}, 29,
  82--97.

\bibitem[{Hochreiter and Schmidhuber(1997)}]{hochreiter1997long}
Hochreiter, S. and Schmidhuber, J. (1997), \enquote{Long short-term memory,}
  \textit{Neural computation}, 9, 1735--1780.

\bibitem[{Jaeger(2007)}]{jaeger2007discovering}
Jaeger, H. (2007), \enquote{Discovering multiscale dynamical features with
  hierarchical echo state networks,} Tech. rep., Jacobs University Bremen.

\bibitem[{Karpatne et~al.(2017)Karpatne, Atluri, Faghmous, Steinbach, Banerjee,
  Ganguly, Shekhar, Samatova, and Kumar}]{karpatne2017theory}
Karpatne, A., Atluri, G., Faghmous, J.~H., Steinbach, M., Banerjee, A.,
  Ganguly, A., Shekhar, S., Samatova, N., and Kumar, V. (2017),
  \enquote{Theory-guided data science: A new paradigm for scientific discovery
  from data,} \textit{IEEE Transactions on Knowledge and Data Engineering}, 29,
  2318--2331.

\bibitem[{Keren and Schuller(2016)}]{keren2016convolutional}
Keren, G. and Schuller, B. (2016), \enquote{Convolutional RNN: an enhanced
  model for extracting features from sequential data,} in \textit{Neural
  Networks (IJCNN), 2016 International Joint Conference on}, IEEE, pp.
  3412--3419.

\bibitem[{Leeds et~al.(2014)Leeds, Wikle, and Fiechter}]{leeds2014emulator}
Leeds, W.~B., Wikle, C.~K., and Fiechter, J. (2014), \enquote{Emulator-assisted
  reduced-rank ecological data assimilation for nonlinear multivariate
  dynamical spatio-temporal processes,} \textit{Statistical Methodology}, 17,
  126--138.

\bibitem[{Luko{\v{s}}evi{\v{c}}ius and
  Jaeger(2009)}]{lukovsevivcius2009reservoir}
Luko{\v{s}}evi{\v{c}}ius, M. and Jaeger, H. (2009), \enquote{Reservoir
  computing approaches to recurrent neural network training,} \textit{Computer
  Science Review}, 3, 127--149.

\bibitem[{Ma et~al.(2017)Ma, Shen, and Cottrell}]{ma2017deep}
Ma, Q., Shen, L., and Cottrell, G.~W. (2017), \enquote{Deep-ESN: A Multiple
  Projection-encoding Hierarchical Reservoir Computing Framework,}
  \textit{arXiv preprint arXiv:1711.05255}.

\bibitem[{MacKay(1992)}]{mackay1992practical}
MacKay, D.~J. (1992), \enquote{A practical Bayesian framework for
  backpropagation networks,} \textit{Neural computation}, 4, 448--472.

\bibitem[{McDermott and Wikle(2017{\natexlab{a}})}]{mcdermott2017bayesian}
McDermott, P.~L. and Wikle, C.~K. (2017{\natexlab{a}}), \enquote{Bayesian
  Recurrent Neural Network Models for Forecasting and Quantifying Uncertainty
  in Spatial-Temporal Data,} \textit{arXiv preprint arXiv:1711.00636}.

\bibitem[{McDermott and Wikle(2017{\natexlab{b}})}]{mcdermott2017ensemble}
--- (2017{\natexlab{b}}), \enquote{An Ensemble Quadratic Echo State Network for
  Nonlinear Spatio-Temporal Forecasting,} \textit{STAT}, 6, 315--330.

\bibitem[{McDermott and Wikle(2018)}]{mcdermott2018deep}
--- (2018), \enquote{Deep echo state networks with uncertainty quantification
  for spatio-temporal forecasting,} \textit{Environmetrics}, e2553.

\bibitem[{Neal(1996)}]{neal1996bayesian}
Neal, R.~M. (1996), \textit{Bayesian learning for neural networks}, New York,
  NY: Springer-Verlag.

\bibitem[{Polson et~al.(2017)Polson, Sokolov, et~al.}]{polson2017deepB}
Polson, N.~G., Sokolov, V., et~al. (2017), \enquote{Deep learning: A bayesian
  perspective,} \textit{Bayesian Analysis}, 12, 1275--1304.

\bibitem[{Polson and Sokolov(2017)}]{polson2017deep}
Polson, N.~G. and Sokolov, V.~O. (2017), \enquote{Deep learning for short-term
  traffic flow prediction,} \textit{Transportation Research Part C: Emerging
  Technologies}, 79, 1--17.

\bibitem[{Quiroz et~al.(2018)Quiroz, Nott, and Kohn}]{quiroz2018gaussian}
Quiroz, M., Nott, D.~J., and Kohn, R. (2018), \enquote{Gaussian variational
  approximation for high-dimensional state space models,} \textit{arXiv
  preprint arXiv:1801.07873}.

\bibitem[{Rasmussen and Williams(2006)}]{rasmussen2006gaussian}
Rasmussen, C.~E. and Williams, C.~K. (2006), \textit{Gaussian processes for
  machine learning}, Cambridge, MA: MIT press.

\bibitem[{Shalev-Shwartz et~al.(2017)Shalev-Shwartz, Shamir, and
  Shammah}]{shalev2017failures}
Shalev-Shwartz, S., Shamir, O., and Shammah, S. (2017), \enquote{Failures of
  deep learning,} \textit{arXiv preprint arXiv:1703.07950}.

\bibitem[{Sheng et~al.(2013)Sheng, Zhao, Wang, and Leung}]{sheng2013prediction}
Sheng, C., Zhao, J., Wang, W., and Leung, H. (2013), \enquote{Prediction
  intervals for a noisy nonlinear time series based on a bootstrapping
  reservoir computing network ensemble,} \textit{IEEE Transactions on neural
  networks and learning systems}, 24, 1036--1048.

\bibitem[{Silver et~al.(2016)Silver, Huang, Maddison, Guez, Sifre, Van
  Den~Driessche, Schrittwieser, Antonoglou, Panneershelvam, Lanctot,
  et~al.}]{silver2016mastering}
Silver, D., Huang, A., Maddison, C.~J., Guez, A., Sifre, L., Van Den~Driessche,
  G., Schrittwieser, J., Antonoglou, I., Panneershelvam, V., Lanctot, M.,
  et~al. (2016), \enquote{Mastering the game of Go with deep neural networks
  and tree search,} \textit{nature}, 529, 484.

\bibitem[{Silver et~al.(2018)Silver, Hubert, Schrittwieser, Antonoglou, Lai,
  Guez, Lanctot, Sifre, Kumaran, Graepel, et~al.}]{silver2018general}
Silver, D., Hubert, T., Schrittwieser, J., Antonoglou, I., Lai, M., Guez, A.,
  Lanctot, M., Sifre, L., Kumaran, D., Graepel, T., et~al. (2018), \enquote{A
  general reinforcement learning algorithm that masters chess, shogi, and Go
  through self-play,} \textit{Science}, 362, 1140--1144.

\bibitem[{Snoek et~al.(2015)Snoek, Rippel, Swersky, Kiros, Satish, Sundaram,
  Patwary, Prabhat, and Adams}]{snoek2015scalable}
Snoek, J., Rippel, O., Swersky, K., Kiros, R., Satish, N., Sundaram, N.,
  Patwary, M., Prabhat, M., and Adams, R. (2015), \enquote{Scalable bayesian
  optimization using deep neural networks,} in \textit{International Conference
  on Machine Learning}, pp. 2171--2180.

\bibitem[{Takens(1981)}]{takens1981detecting}
Takens, F. (1981), \enquote{Detecting strange attractors in turbulence,}
  \textit{Lecture notes in mathematics}, 898, 366--381.

\bibitem[{Tobler(1970)}]{tobler1970computer}
Tobler, W.~R. (1970), \enquote{A computer movie simulating urban growth in the
  Detroit region,} \textit{Economic geography}, 46, 234--240.

\bibitem[{Tong and Tanaka(2018)}]{tong2018reservoir}
Tong, Z. and Tanaka, G. (2018), \enquote{Reservoir Computing with Untrained
  Convolutional Neural Networks for Image Recognition,} in \textit{2018 24th
  International Conference on Pattern Recognition (ICPR)}, IEEE, pp.
  1289--1294.

\bibitem[{Tran et~al.(2018)Tran, Nguyen, Nott, and Kohn}]{tran2018bayesian}
Tran, M.-N., Nguyen, N., Nott, D., and Kohn, R. (2018), \enquote{Bayesian Deep
  Net GLM and GLMM,} \textit{arXiv preprint arXiv:1805.10157}.

\bibitem[{Triefenbach et~al.(2013)Triefenbach, Jalalvand, Demuynck, and
  Martens}]{triefenbach2013acoustic}
Triefenbach, F., Jalalvand, A., Demuynck, K., and Martens, J. (2013),
  \enquote{Acoustic modeling with hierarchical reservoirs,} \textit{IEEE
  Transactions on Audio, Speech, and Language Processing}, 21, 2439--2450.

\bibitem[{Wikle et~al.(2019)Wikle, Zammit-Mangion, and
  Cressie}]{wikle2019spatio}
Wikle, C., Zammit-Mangion, A., and Cressie, N. (2019), \textit{Spatio-Temporal
  Statistics with R}, Boca Raton, FL: Chapman and Hall/CRC.

\bibitem[{Wikle et~al.(1998)Wikle, Berliner, and
  Cressie}]{wikle1998hierarchical}
Wikle, C.~K., Berliner, L.~M., and Cressie, N. (1998), \enquote{Hierarchical
  Bayesian space-time models,} \textit{Environmental and Ecological
  Statistics}, 5, 117--154.

\bibitem[{Wikle and Hooten(2010)}]{wikle2010general}
Wikle, C.~K. and Hooten, M.~B. (2010), \enquote{A general science-based
  framework for dynamical spatio-temporal models,} \textit{Test}, 19, 417--451.

\bibitem[{Wikle et~al.(2001)Wikle, Milliff, Nychka, and
  Berliner}]{wikle2001spatiotemporal}
Wikle, C.~K., Milliff, R.~F., Nychka, D., and Berliner, L.~M. (2001),
  \enquote{Spatiotemporal hierarchical Bayesian modeling tropical ocean surface
  winds,} \textit{Journal of the American Statistical Association}, 96,
  382--397.

\bibitem[{Xingjian et~al.(2015)Xingjian, Chen, Wang, Yeung, Wong, and
  Woo}]{xingjian2015convolutional}
Xingjian, S., Chen, Z., Wang, H., Yeung, D.-Y., Wong, W.-K., and Woo, W.-c.
  (2015), \enquote{Convolutional LSTM network: A machine learning approach for
  precipitation nowcasting,} in \textit{Advances in neural information
  processing systems}, pp. 802--810.

\end{thebibliography}
\bibliographystyle{asa}

\end{document}